\documentclass[sigconf, nonacm]{acmart}
\usepackage{balance}
\usepackage{graphicx}
\usepackage{multirow}
\AtBeginDocument{%
  }

\setcopyright{cc}
\setcctype{by}

\begin{document}

\title{Stationarity-Aware Retrieval-Augmented Time Series Forecasting}

\author{Shiqiao Zhou}
\orcid{0009-0003-3515-2864}
\affiliation{%
  \institution{University of Birmingham}
  \city{Birmingham}
  \country{United Kingdom}}
\email{sxz363@student.bham.ac.uk}

\author{Holger Schöner}
\orcid{0009-0009-9341-538X}
\affiliation{%
  \institution{Siemens AG}
  \city{Munich}
  \country{Germany}}
\email{holger.schoener@siemens.com}

\author{Zipeng Wu}
\orcid{0000-0002-1171-8187}
\affiliation{%
  \institution{University of Birmingham}
  \city{Birmingham}
  \country{United Kingdom}}
\email{zxw365@student.bham.ac.uk}

\author{Edouard Fouché}
\authornote{The contribution was carried out while this co-author was employed by Siemens AG.}
\orcid{0000-0003-0157-7648}
\affiliation{%
  \institution{Siemens AG}
  \city{Munich}
  \country{Germany}}
\email{edouard.fouche@siemens-healthineers.com}

\author{IAG Wilson}
\orcid{0000-0003-4083-597X}
\affiliation{%
  \institution{University of Birmingham}
  \city{Birmingham}
  \country{United Kingdom}}
\email{i.a.g.wilson@bham.ac.uk}

\author{Shuo Wang}
\authornote{Corresponding author}
\orcid{0000-0003-1380-6428}
\affiliation{%
  \institution{University of Birmingham}
  \city{Birmingham}
  \country{United Kingdom}}
\email{s.wang.2@bham.ac.uk}

\thanks{© {Shiqiao Zhou, Holger Schöner, Zipeng Wu, Edouard Fouché, IAG Wilson, and Shuo Wang | ACM} {2026}. This is the authors' accepted version of a paper accepted to KDD 2026. The definitive Version of Record will be published at: http://doi.org/10.1145/3770855.3817813.}

\renewcommand{\shortauthors}{Shiqiao Zhou et al.}

\begin{abstract}
  Time series forecasting relies on historical patterns, but real-world series often exhibit non-stationarity and regime shifts that challenge fully parametric forecasters. Inspired by Retrieval-Augmented Generation (RAG), recent work augments forecasters by retrieving relevant historical segments and using them as external evidence at inference time. However, due to the intrinsic non-stationarity of real-world time series, a highly similar past segment does not necessarily imply a similar future, rendering similarity-only retrieval brittle and prone to redundancy. We propose \textbf{S}tationarity-\textbf{A}ware \textbf{R}etrieval-\textbf{A}ugmented Time Series \textbf{F}orecasting (SARAF), a framework that adaptively balances relevance and diversity in retrieval. SARAF first forms a candidate pool via temporal similarity with time-aligned enhancement, then applies a diversity-aware selection strategy to cover heterogeneous historical regimes, with the diversification strength automatically modulated by dataset-level stationarity. Moreover, SARAF uses stationarity-aware aggregation to fuse the retrieved futures. Extensive experiments on eight real-world datasets show that SARAF achieves competitive forecasting performance and improves average accuracy and robustness over strong baselines, with particularly clear benefits under challenging non-stationary settings. Code: \url{https://github.com/ShiqiaoZhou/SARAF}.
\end{abstract}



\maketitle

\section{Introduction}
Time series forecasting is a longstanding problem with broad impact in domains such as traffic~\cite{lippi2013short}, energy~\cite{daut2017building}, finance~\cite{poon2003forecasting}, and climate~\cite{price2025probabilistic}. Given a historical context, the goal is to predict future values over a specified horizon. Over the past decades, forecasting methods have evolved from classical statistical models such as ARIMA~\cite{box2015time} to modern deep learning approaches that better capture nonlinearity and multivariate dependencies. Modern forecasters range from lightweight architectures, often equipped with normalization and decomposition for efficiency and strong performance~\cite{zeng2023transformers, lin2024cyclenet}, to attention-based models that learn expressive temporal representations and better capture long-range dependencies under complex multivariate dynamics~\cite{nietime, liu2022non, wu2021autoformer, liuitransformer}. Furthermore, time series foundation models~\cite{das2024decoder, ansari2024chronos, woo2024unified} have recently emerged as a unified pretraining-based backbone, learning transferable representations from large-scale time series corpora and improving generalization across heterogeneous and non-stationary settings.

\begin{figure}[t]
    \centering
    \includegraphics[width=1\linewidth]{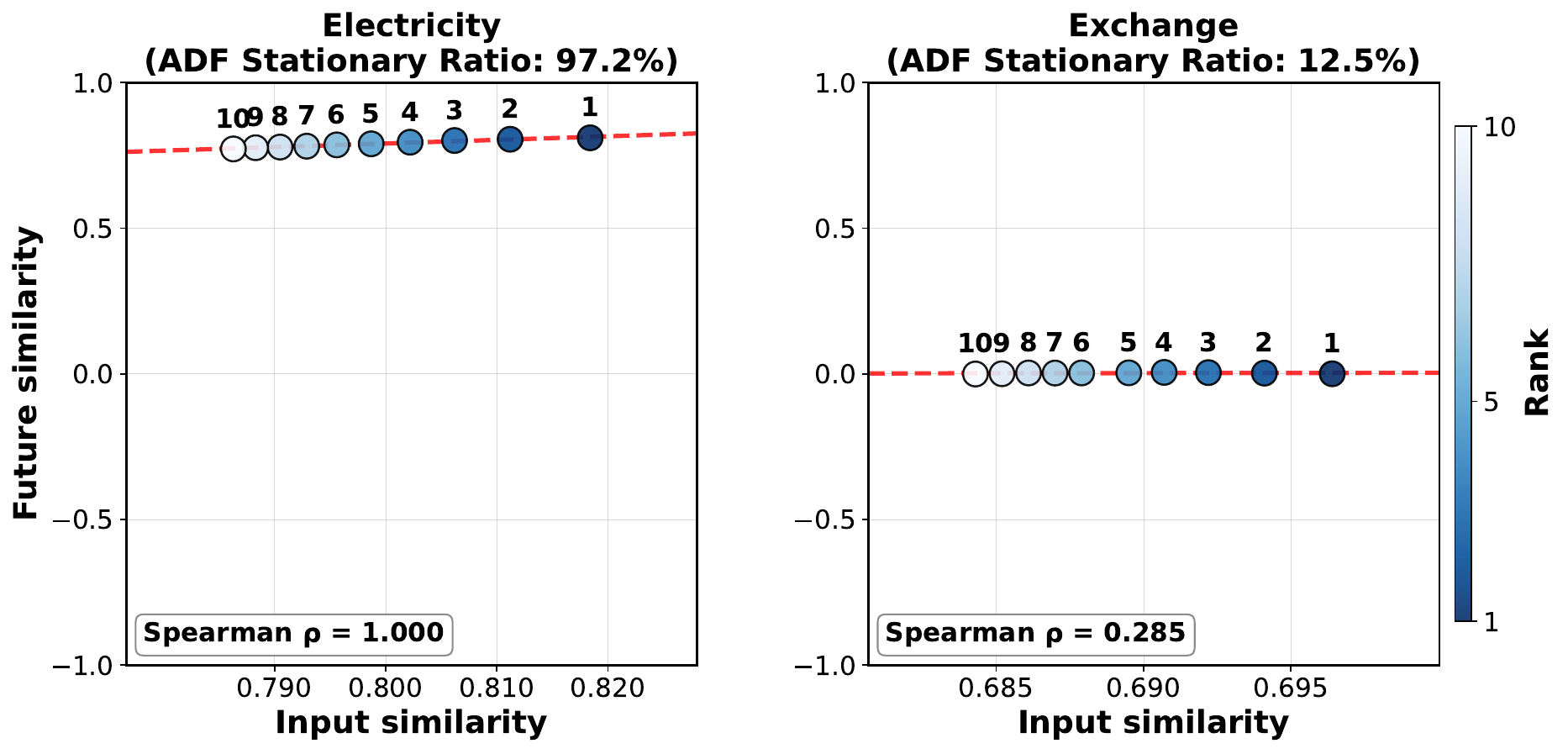}
    \caption{An example motivating SARAF: Under the 720-input and 96-horizon setting, the x-axis shows the Pearson correlation similarity between a test input and its retrieved training inputs (ranked by input similarity), while the y-axis shows the Pearson correlation similarity between the corresponding retrieved futures and the ground-truth future. Comparing a more stationary dataset (Electricity) with a non-stationary one (Exchange) shows that higher input-similarity ranks do not necessarily correspond to higher future-similarity ranks; some retrieved futures show near-zero similarity to the ground truth, suggesting that similarity-only retrieval can be brittle under non-stationarity. More examples can be found in Appendix ~\ref{Motivation_app}.}
    \label{fig:motivation} 
\end{figure}

Motivated by the success of Retrieval-Augmented Generation (RAG) in large language models~\cite{lewis2020retrieval}, recent works~\cite{hanretrieval, ning2025ts, chen2025trace} have explored retrieval-augmented forecasting, a paradigm that explicitly retrieves historically similar segments as external evidence and integrates them into the forecasting pipeline. Such a retrieval mechanism can be incorporated during training and testing, enabling the model to learn how to exploit retrieved patterns while retaining the flexibility to condition predictions on instance-specific references. By retrieving relevant historical cases, the forecaster can ground its prediction on concrete continuations instead of relying only on patterns absorbed during training. This is particularly beneficial for long-tail behaviors: loss-minimization tends to prioritize frequent regimes, so rare yet consequential motifs are often under-emphasized by fully parametric forecasters, whereas retrieval gives the model an additional chance to fetch a few closely related historical segments as explicit references.

However, most existing retrieval-augmented forecasters implicitly assume that \emph{``similar pasts imply similar futures''}, and thus perform similarity-only top-$K$ retrieval based on the input history.
To probe this assumption, we conduct a simple diagnostic experiment under the 720-input and 96-horizon setting, as shown in Fig.~\ref{fig:motivation}.
Specifically, we choose two widely used time series forecasting benchmarks, Electricity and Exchange, as representative cases for this diagnosis.
To demonstrate the relevance of stationarity, we report the ADF (Augmented Dickey--Fuller) stationarity ratio~\cite{dickey1979distribution}, defined as the fraction of channels whose ADF test rejects the unit-root null. Under this definition, 97.2\% of channels in Electricity are classified as stationary, while only 12.5\% in Exchange are stationary.
For each test instance, we compute Pearson
correlation similarity between its input segment and all candidate segments in the training set, rank them, and retrieve the top-10 nearest neighbors. 
We then compute the Pearson correlation similarity, between each retrieved neighbor's future segment and the test instance's ground-truth future, and aggregate the results across test samples to obtain the rank-wise statistics in the figure. 
Importantly, Fig.~\ref{fig:motivation} reports the Spearman rank correlation $\rho$ between the input-similarity ranking used for retrieval and the resulting future-similarity ranking that reflects the usefulness of retrieved futures.

From Fig.~\ref{fig:motivation}, for Electricity, $\rho=1.000$ indicates an almost perfectly rank-consistent behavior: neighbors that are more similar in the input space also tend to yield futures that are consistently positive and relatively high in similarity to the ground-truth future, making similarity-based top-$K$ retrieval largely reliable. 
In contrast, for Exchange, $\rho=0.285$ means it still reveals a clear rank mismatch in practice: input similarity is not a dependable proxy for future similarity, and many retrieved futures have little to no similarity to the ground truth. 

This sharp contrast suggests that similarity-only retrieval can become brittle on highly non-stationary datasets where ``similar pasts'' may correspond to markedly different futures.
Such brittleness often comes with a practical issue in sliding-window databases: the retrieved Top-$K$ results are frequently highly redundant. In particular, redundant retrieval wastes the limited Top-$K$ budget by returning near-duplicate evidence, reducing the effective information content. More importantly, under non-stationarity, repeated but mismatched cases can be aggregated into a misleading ``consensus'', amplifying errors and even yielding severely degraded retrieval ranks and unstable forecasts. In practice, many retrieval-augmented forecasters ~\cite{ning2025ts,hanretrieval, du2025predicting, liu2025improving} adopt a simple safeguard: when retrieved evidence appears unreliable, they down-weight it to reduce its influence during training and inference.
However, such attenuation has two key limitations: (i) it does not resolve the root cause---the mismatch between input-based retrieval and future relevance---and (ii) it can overly suppress partially useful neighbors, effectively reverting the model to a purely parametric forecaster and diminishing retrieval's benefits on long-tail or regime-specific patterns.

To this end, we propose \textbf{SARAF}, a \textbf{s}tationarity-\textbf{a}ware \textbf{r}etrieval-\textbf{a}ugmented \textbf{f}orecasting framework that improves evidence quality under distribution shifts and heterogeneous regimes. 
SARAF is built upon two key ideas. First, we introduce a \emph{time-aligned retrieval enhancement} that explicitly favors candidates from historically aligned periods (e.g., the same hour-of-day or day-of-week). This provides an additional matching axis beyond morphology, reducing ``morphologically similar but temporally mismatched'' neighbors that often yield inconsistent futures. Notably, this cue is broadly applicable across both stationary and non-stationary datasets, as temporal alignment helps recover truly comparable contexts. 

Second, SARAF makes retrieval \emph{stationarity-adaptive} in both selection and fusion. We introduce a simple stationarity estimation method.
On low-stationarity datasets, where similarity-only retrieval is brittle, SARAF strengthens diversity-aware selection to cover heterogeneous regimes and mitigate redundancy, and uses a smoother fusion that distributes non-trivial weights across more retrieved futures for robustness.
On high-stationarity datasets, SARAF prioritizes the most similar neighbors and applies a sharper aggregation that concentrates weight on top-ranked futures to preserve precision. Finally, we adopt a lightweight fusion by simply averaging the retrieval results with the naive prediction, allowing the whole pipeline to directly benefit from the retrieved evidence.

Our main contributions are:
\begin{itemize}
    \item We empirically demonstrate that the reliability of similarity-based retrieval is strongly dataset-dependent: future similarity between retrieved results and the ground truth varies substantially across stationarity levels.
    \item We propose SARAF, a stationarity-aware retrieval-augmented forecasting framework that combines (i) time-aligned retrieval enhancement and (ii) stationarity-controlled diversity-based retrieval with adaptive fusion, guided by a simple dataset-level stationarity estimator.
    \item We conduct extensive experiments on eight datasets, showing that SARAF achieves competitive forecasting performance against strong baselines, with useful gains and robust behavior across datasets with different stationarity levels.
\end{itemize}

\section{Related Work}
\subsection{Time Series Forecasting}
Time series forecasting has evolved from classical statistical models to deep representation learning.
Traditional methods such as ARIMA \cite{box2015time} assume stationarity and are interpretable, but often struggle with nonlinear dynamics and high-dimensional dependencies.
Deep learning mitigated these limitations with recurrent models such as LSTMs \cite{hochreiter1997long} and probabilistic frameworks like DeepAR \cite{salinas2020deepar}.
Transformers further improved long-range modeling, with efficiency-oriented designs such as Informer \cite{zhou2021informer} and Autoformer \cite{wu2021autoformer} reducing attention cost and injecting inductive biases.
Meanwhile, the community has revisited architectural complexity: DLinear \cite{zeng2023transformers} shows that simple normalized linear mappings can be highly competitive, motivating refined representations such as PatchTST \cite{nie2022time} with patch-wise tokenization.
Beyond the time domain, frequency and periodic modeling remain active, where TimesNet \cite{wu2022timesnet}, CycleNet \cite{lin2024cyclenet}, and MoFo \cite{ma2025mofo} explicitly encode multi-periodicity under heterogeneous dynamics.
Real-world series also exhibit strong non-stationarity with distribution and concept drifts. RevIN \cite{kim2021reversible} provides a lightweight normalization remedy, while Non-stationary Transformers \cite{liu2022non} and SAN \cite{liu2023adaptive} incorporate adaptivity through de-stationary attention and dynamic normalization to preserve informative non-stationary cues.
In contrast, TimeBridge \cite{liu2024timebridge} cautions that overly aggressive stationarization may erase long-term multivariate structure, and proposes dual-scale representation learning.

Nevertheless, most forecasting models are optimized for frequent, dominant patterns, which can leave rare or emerging regimes in the long tail under-modeled. Retrieval-augmented forecasting alleviates this by augmenting the current context with retrieved historical analogues, injecting instance-specific evidence that improves robustness beyond what static model parameters alone can capture.

\subsection{Retrieval-Augmented Time Series Forecasting}
With the rise of RAG in large language models, the time series forecasting community has increasingly explored retrieval-based mechanisms to improve long-horizon prediction. Existing methods can be broadly grouped into two categories.

\textbf{Retrieval as Input Augmentation.} A common line of work retrieves historical segments and feeds them to the forecasting model together with the query context. For example, \textsc{TimeRAG}~\cite{zhang2025timeraf} clusters time series segments and retrieves similar instances using DTW, which are then provided to an LLM to generate forecasts. \textsc{RAFT}~\cite{hanretrieval} performs retrieval via multi-period matching and Pearson correlation, and leverages the retrieved \emph{future} segments to augment the original input for prediction. More recent approaches, such as \textsc{RAF}~\cite{li2025retrieval, tire2024retrieval}, \textsc{TRACE}~\cite{chen2025trace}, and \textsc{TS-RAG}~\cite{ning2025ts}, integrate retrieval with time series foundation models to better exploit large-scale pretraining and improve generalization.

\textbf{Retrieval as Post-hoc Refinement.} Another stream treats retrieval as a lightweight, model-agnostic module attached to an existing forecaster to correct or ensemble predictions. \textsc{PFRP}~\cite{du2025predicting} constructs a global memory bank that stores categorized historical patterns, and fuses the base forecaster's prediction with an additional prediction generated from the memory. \textsc{PIR}~\cite{liu2025improving} instead performs post-hoc correction: after the base model produces an initial forecast, it retrieves similar training segments and uses them to refine the prediction.

Despite their effectiveness, most retrieval-augmented forecasters still rely on pure similarity-based retrieval, which can be brittle on non-stationary datasets: histories that look similar in the past may yield futures that diverge sharply from the ground truth. Prior work often addresses this by down-weighting retrieval when evidence appears unreliable. In contrast, SARAF strengthens retrieval itself under non-stationarity by selecting a stationarity-controlled, diverse subset of candidates, and directly fusing them with the backbone prediction rather than suppressing the retrieval branch.

\section{Methodology}
\label{sec:method}

\subsection{Problem Formulation}
\label{sec:problem}
We formulate multivariate time series forecasting as learning a predictive function $f(\cdot)$ that maps a historical window to a future horizon. Specifically, given a past input sequence $\mathbf{X}_{t-L+1:t}=\{x_{t-L+1},\dots,x_t\}\in\mathbb{R}^{L\times C}$, where $L$ denotes the look-back window and $C$ denotes the number of channels, the objective is to train $f$ to produce the next-$H$-step forecast: $\hat{\mathbf{Y}}_{t+1:t+H}=f(\mathbf{X}_{t-L+1:t}) \in \mathbb{R}^{H\times C}$, 
which approximates the ground-truth future sequence $\mathbf{Y}_{t+1:t+H}=\{y_{t+1},\dots,y_{t+H}\}$. In practice, $f$ is learned by minimizing a forecasting loss between $\hat{\mathbf{Y}}_{t+1:t+H}$ and $\mathbf{Y}_{t+1:t+H}$ over the training set.

\begin{figure*}[h]
  \centering
  \includegraphics[width=\linewidth, trim = 0.8cm 3.7cm 0.8cm 0.4cm, clip]{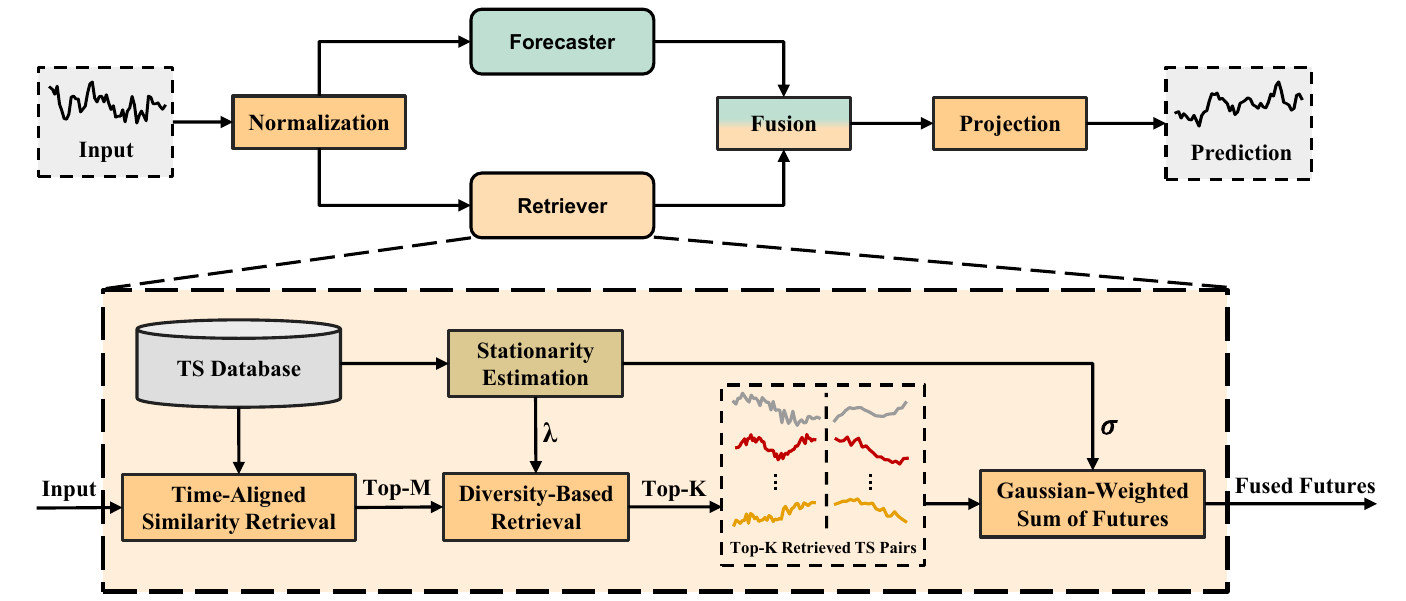}
  \caption{Model architecture. SARAF augments a forecaster with a stationarity-aware retriever. Given an input window, SARAF first performs time-aligned similarity retrieval to form a Top-$M$ candidate pool, then applies a stationarity-guided diversity selection to obtain Top-$K$ retrieved pairs. The corresponding futures are aggregated via a stationarity-conditioned Gaussian weighting to produce retrieval-based prediction, which is then combined with the backbone prediction through a lightweight fusion and projection for the final forecast.}
  \label{SARAF}
\end{figure*}

\subsection{Overview}
\label{sec:overview}

As shown in Figure~\ref{SARAF}, SARAF first normalizes the input query $\mathbf{x}_q$ and computes its similarity to all historical windows in a time series (TS) database constructed from the training set. Based on these time-aligned TS similarities, we form a candidate set by selecting the top-$M$ most similar time series pairs $\{(\mathbf{x}_i,\mathbf{y}_i)\}_{i=1}^{M}$. We then perform diversity-based retrieval over these $M$ candidates to select $K$ pairs $(M \gg K)$, where the diversification strength is controlled by a dataset-level stationarity score $\bar{s}$ estimated from the TS database. Given the retrieved $\{(\mathbf{x}_i,\mathbf{y}_i)\}_{i=1}^{K}$ pairs, we obtain a fused retrieval-based forecast via a Gaussian-weighted aggregation of the retrieved futures, where the kernel bandwidth $\sigma$ is also modulated by $\bar{s}$.
Finally, we average the retrieved futures with the naive prediction of the forecaster, and feed the fused representation into an output projection layer to produce the forecast. Overall, SARAF couples a time-aligned retrieval enhancement with a stationarity-aware controller that modulates both retrieval diversity and aggregation sharpness. Together, these components improve the reliability of retrieved evidence across heterogeneous regimes and yield more robust forecasts under non-stationarity.

\subsection{Time Series Database}
\label{sec:stationarity}
\subsubsection{Time Series Database Construction}
To construct the time series (TS) database, we use a sliding window with stride = 1 to extract historical segments of length $L$ only from the training set. Each segment is paired with its corresponding future sequence of length $H$. We denote the resulting database of $N$ time series pairs as $\mathcal{D}_{\mathrm{TS}}=\{(\mathbf{x}_i,\mathbf{y}_i)\}_{i=1}^{N}$, where $\mathbf{x}_i\in\mathbb{R}^{L\times C}$ and $\mathbf{y}_i\in\mathbb{R}^{H\times C}$.

\subsubsection{Stationarity Estimation}
    To estimate dataset-level stationarity, we sample historical segments from $\mathcal{D}_{\mathrm{TS}}$, compute an instance-wise stationarity score for each segment, and take their average as a proxy for the overall stationarity of the training data. This estimation is computationally efficient for multivariate TS databases and directly provides a stationarity score for subsequent retrieval steps.
In contrast, outputs from classical tests such as ADF~\cite{dickey1979distribution} and KPSS~\cite{shin1992kpss} are univariate and more expensive to compute at scale, which makes them less practical for stationarity control in our retrieval pipeline.

Specifically, for a sample $\mathbf{x}_i\in\mathbb{R}^{L\times C}$, we partition it into $W$ non-overlapping sub-windows. For each sub-window, we compute the per-channel local mean and standard deviation, and quantify their \emph{variation} across sub-windows. Concretely, let $v_\mu$ and $v_\sigma$ denote the standard deviation (over sub-windows) of the sub-window means and standard deviations, respectively. To make the score scale-invariant, we normalize these variations by a global scale term, defined as the standard deviation of the dataset $\{\mathbf{x}_i\}_{i=1}^{N}$ and denoted by $\bar{\sigma}$. The stationarity score for sample $\mathbf{x}_i$ is defined as
\begin{equation}
    \tilde{s}_i=\frac{1}{2}\left(\left[1-\min\!\left(1,\frac{v_\mu}{\bar{\sigma}}\right)\right]
    +\left[1-\min\!\left(1,\frac{v_\sigma}{\bar{\sigma}}\right)\right]\right)\in[0,1],
    \label{eq:stationarity_score}
\end{equation}
where larger values indicate that the local mean and variance are more stable over time. Finally, we estimate dataset-level stationarity by averaging the sample-level scores over all $N$ samples:
\begin{equation}
    \bar{s}=\frac{1}{N}\sum_{i=1}^{N}\tilde{s}_i.
    \label{eq:dataset_stationarity}
\end{equation}
The computed stationarity scores are all reported in Appendix~\ref{app:stationarity_score}.

\subsection{Retriever Design}
Building on conventional similarity-based retrieval, we design the retriever in three steps.
First, we add a time-aligned enhancement that favors candidates from temporally aligned periods, increasing the chance of retrieving corresponding cycles.
Second, we perform diversity-aware selection on top of similarity ranking to reduce redundancy and better handle datasets with different degrees of non-stationarity.
Finally, we aggregate the Top-$K$ retrieved futures with Gaussian-kernel weights to form a robust fused retrieval forecast.

\subsubsection{Time-Aligned Retrieval Enhancement}
Time series commonly exhibit calendar-driven regularities (e.g., diurnal, weekly, and seasonal patterns). Purely pattern-based retrieval may therefore return segments that are morphologically similar yet misaligned in temporal context. To inject this inductive bias, we compute a \emph{time-aligned bonus} that rewards candidates whose timestamps align with the query at multiple temporal resolutions.

Given a query batch with timestamps $\{\mathbf{t}_{q}\}_{q=1}^{B}$ and training timestamps $\{\mathbf{t}_i\}_{i=1}^{N}$, we compute a bonus matrix $\mathbf{B}\in\mathbb{R}^{B\times N}$ as
\begin{equation}
\mathbf{B}_{q,i}=\sum_{r\in\mathcal{R}}\lambda_r\,\phi_r(\mathbf{t}_{q},\mathbf{t}_i),
\label{eq:time_bonus_general}
\end{equation}
where $\mathcal{R}$ includes hour-of-day, day-of-week, month-of-year, and minute-of-hour when available. 
For cyclic components (hour, month, and optionally minute), we compute their discrepancy using a circular distance to account for wrap-around (e.g., 23 and 0 are close),
\[
d_r=\min\!\left(|t^{(r)}_{q}-t^{(r)}_{i}|,\;P_r-|t^{(r)}_{q}-t^{(r)}_{i}|\right),
\]
and implement $\phi_r(\cdot,\cdot)$ via an exponential kernel on $d_r$, where $P_r$ is the period (24 for hours, 12 for months, and 60 for minutes). 
For the weekly component, we assign the highest bonus to exact weekday matches, and a smaller bonus when the query and candidate share the same coarse type (weekday versus weekend). 
We use comparable weights across temporal components to balance daily, weekly, and seasonal cues so that no single granularity dominates the bonus. 
Finally, we rescale $\mathbf{B}$ by its maximum entry so that all bonus values lie in $[0,1]$.

\begin{table*}[t]
\centering
\small
\caption{Dataset statistics and stationarity scores. }
\label{tab:dataset_stats_simple}
\begin{tabular}{c c c c c c c c c}
\toprule
Dataset & Channel & Input Length & Forecasting Horizon & Dataset Size & Split Ratio & Frequency & Stationarity Score & ADF Stationary Ratio \\
\midrule
ETTh1        & 7   & 720 & \{96, 192, 336, 720\} & 14,440 & 6/2/2 & 1 hour  & 0.7041 & 100.0\% \\
ETTh2        & 7   & 720 & \{96, 192, 336, 720\} & 14,440 & 6/2/2 & 1 hour  & 0.5731 & 57.1\%  \\
ETTm1        & 7   & 720 & \{96, 192, 336, 720\} & 57,600 & 6/2/2 & 15 min  & 0.6466 & 100.0\% \\
ETTm2        & 7   & 720 & \{96, 192, 336, 720\} & 57,600 & 6/2/2 & 15 min  & 0.6080 & 85.7\%  \\
Exchange     & 8   & 720 & \{96, 192, 336, 720\} & 7,588  & 7/1/2 & 1 day  & 0.4203 & 12.5\%  \\
Solar & 137 & 720 & \{96, 192, 336, 720\} & 52,560 & 7/1/2 & 10 min  & 0.7439 & 100.0\% \\
Electricity  & 321 & 720  & \{96, 192, 336, 720\} & 26,304 & 7/1/2 & 1 hour  & 0.8648 & 97.2\%  \\
Traffic      & 862 & 720 & \{96, 192, 336, 720\} & 17,544 & 7/1/2 & 1 hour  & 0.8628 & 100.0\% \\
\bottomrule
\end{tabular}
\end{table*}

\subsubsection{Time-Aligned Similarity Retrieval}
With the time-aligned bonus $\mathbf{B}$ calculated, we next perform similarity-based retrieval to identify relevant historical segments. 
Given a query window $\mathbf{x}_q\in\mathbb{R}^{L\times C}$ and a candidate training window $\mathbf{x}_i$, we adopt Pearson correlation as the similarity function, as it is robust to scale variations and mean shifts while emphasizing co-movement trends~\cite{hanretrieval}. We compute the similarity on vectorized, centered windows:
\begin{equation}
\mathbf{S}_{\text{temporal}}(q,i)
= \frac{\left\langle \mathrm{vec}(\tilde{\mathbf{x}}_q), \mathrm{vec}(\tilde{\mathbf{x}}_i)\right\rangle}{
\|\mathrm{vec}(\tilde{\mathbf{x}}_q)\|_2\cdot \|\mathrm{vec}(\tilde{\mathbf{x}}_i)\|_2},
\label{eq:similarity}
\end{equation}
where $\tilde{\mathbf{x}}$ denotes the centered window and $\mathrm{vec}(\cdot)$ flattens an $L\times C$ matrix into $\mathbb{R}^{LC}$.

We then incorporate the time-aligned bonus by combining $\mathbf{S}_{\text{temporal}}$ with the bonus matrix:
\begin{equation}
\mathbf{S}_{\text{sim}}(q,i) = (1-\alpha_{\text{time}})\,\mathbf{S}_{\text{temporal}}(q,i) + \alpha_{\text{time}}\,\mathbf{B}_{q,i},
\label{eq:time_aware_fusion}
\end{equation}
where $\alpha_{\text{time}}$ controls the contribution of time-aligned enhancement. 
Finally, we rank all candidates by $\mathbf{S}_{\text{sim}}(q,i)$ and retain the Top-$M$ segments as the retrieval candidate pool for subsequent selection.

\subsubsection{Diversity-Based Retrieval}
The initial Top-$M$ pool contains highly relevant candidates under $\mathbf{S}_{\text{sim}}$, yet the \emph{optimal} set of $K$ retrieved windows is inherently \emph{stationarity-dependent}. 
We therefore refine the pool using a stationarity-guided \emph{stochastic} MMR (Maximal Marginal Relevance) procedure, which balances query relevance and diversity among selected windows to better cover plausible regimes, particularly under non-stationarity.

\paragraph{Balance parameter}
Original MMR \cite{carbonell1998use} uses a balance coefficient $\lambda\in[0,1]$ to trade off relevance ($\lambda=1$) and diversity ($\lambda=0$). 
We set $\lambda$ as a dataset-aware function of the dataset-level stationarity $\bar{s}$:
\begin{equation}
    \lambda(\bar{s})
    = \lambda_{\min}+\bar{s}\,(\lambda_{\max}-\lambda_{\min}),
    \label{eq:lambda}
\end{equation}
where $\lambda_{\min}$ and $\lambda_{\max}$ are fixed hyperparameters. 
Thus, less stationary datasets (smaller $\bar{s}$) yield smaller $\lambda$ and encourage stronger diversification.

\paragraph{Stochastic MMR}
Let $\mathcal{C}$ denote the Top-$M$ candidates ranked by $\mathbf{S}_{\text{sim}}(q,i)$ in Eq.~\eqref{eq:time_aware_fusion} (with $M\gg K$), and let $\mathcal{R}$ be the selected set. 
We first include the most similar candidate as an anchor,
$\mathcal{R}\leftarrow\{\arg\max_{\mathbf{x}_i\in\mathcal{C}} \mathbf{S}_{\text{sim}}(q,i)\}$, and then select the remaining $K-1$ windows iteratively.

At each step, for every candidate $\mathbf{x}_i\in\mathcal{C}\setminus\mathcal{R}$, we compute
\begin{equation}
    \mathrm{MMR}(i)
    = \lambda(\bar{s}) \cdot \mathbf{S}_{\text{sim}}(q,i)
    - (1-\lambda(\bar{s}))\cdot \max_{\mathbf{x}_j\in \mathcal{R}} \Delta(i,j),
    \label{eq:mmr_proxy}
\end{equation}
where $\Delta(i,j)$ approximates the redundancy between two candidates. 
To keep selection efficient, we avoid computing full pairwise similarity among the $M$ candidates (an $O(M^2)$ operation) and instead use a lightweight proxy based on query-similarity closeness:
\begin{equation}
    \Delta(i,j) = 1 - \left|\mathbf{S}_{\text{sim}}(q,i)-\mathbf{S}_{\text{sim}}(q,j)\right|.
    \label{eq:redundancy_proxy}
\end{equation}
This proxy discourages selecting candidates that provide highly similar retrieval evidence, encouraging broader coverage when the series is non-stationary.

Finally, rather than deterministically selecting $\arg\max_i \mathrm{MMR}(i)$, we sample the next candidate from a softmax distribution:
\begin{equation}
    p(i)=\mathrm{Softmax}\!\left(\mathrm{MMR}(i)\right).
\end{equation}
We repeat this procedure until $|\mathcal{R}|=K$.

\subsubsection{Adaptive Gaussian Weighting}
Let $\mathcal{R}=\{\mathbf{x}_{r_1},\dots,\mathbf{x}_{r_K}\}$ denote the retrieved input windows. 
We assign a normalized weight $w_k$ to each retrieved item via a Gaussian kernel whose bandwidth is conditioned on the dataset-level stationarity:
\begin{equation}
    \sigma(\bar{s})=\sigma_{\min}+(1-\bar{s})\big(\sigma_{\max}-\sigma_{\min}\big),
    \label{eq:sigma}
\end{equation}
where $\sigma_{\min}$ and $\sigma_{\max}$ are fixed hyperparameters controlling the sharpness range.
The weights are computed as
\begin{equation}
    w_k
    = \frac{\exp\!\left(-\frac{d_k^2}{2\sigma(\bar{s})^2}\right)}{\sum_{j=1}^{K}\exp\!\left(-\frac{d_j^2}{2\sigma(\bar{s})^2}\right)},
    \quad
    d_k = 1-\mathbf{S}_{\text{sim}}(q,r_k).
    \label{eq:gaussian_weight}
\end{equation}

\subsection{Retrieval-Augmented Prediction}
\label{sec:prediction}

\subsubsection{Retrieval-based prediction}
For each retrieved input window $\mathbf{x}_{r_k}$, let $\mathbf{y}_{r_k}\in\mathbb{R}^{H\times C}$ denote its paired future segment.  
We form the retrieval-based prediction by a weighted aggregation:
\begin{equation}
    \hat{\mathbf{Y}}_{\mathrm{ret}} = \sum_{k=1}^{K} w_k\, \mathbf{y}_{r_k}.
    \label{eq:retrieval_pred}
\end{equation}

\subsubsection{Direct prediction}
In parallel, we use a lightweight linear layer as time series forecaster that maps the centered query window to the forecasting horizon:
\begin{equation}
    \hat{\mathbf{Y}}_{\mathrm{direct}} = \mathbf{W}\tilde{\mathbf{x}}_q + \mathbf{b},
    \label{eq:direct_pred}
\end{equation}
where $\mathbf{W}\in\mathbb{R}^{H\times L}$ is applied along the temporal dimension and shared across channels, and $\mathbf{b}\in\mathbb{R}^{H\times C}$ is a learnable bias.

\subsubsection{Fusion and Final Projection}
We keep fusion lightweight by simply averaging the direct forecast and the retrieval-based forecast:
\begin{equation}
    \hat{\mathbf{Y}} = \frac{1}{2}\left(\hat{\mathbf{Y}}_{\mathrm{direct}} + \hat{\mathbf{Y}}_{\mathrm{ret}}\right).
    \label{eq:fusion_avg}
\end{equation} 

Finally, we apply a linear projection along the forecasting horizon to obtain the final prediction:
\begin{equation}
    \hat{\mathbf{Y}}_{\mathrm{final}} = \mathrm{Linear}(\hat{\mathbf{Y}}).
    \label{eq:final_linear}
\end{equation}

\begin{table}[]
\centering
\small
\caption{Ablation study of the retrieval mechanism in SARAF. We report the average MSE and MAE across four forecasting horizons. The best result in each row is highlighted in bold.}
\label{tab:saraf_retrieval_ablation}
\resizebox{\columnwidth}{!}{%
\begin{tabular}{l cc cc cc cc}
\toprule
\textbf{Models}
& \multicolumn{2}{c}{\textbf{SARAF}} 
& \multicolumn{2}{c}{\textbf{w/o Retriever}} 
& \multicolumn{2}{c}{\textbf{w/o Forecaster}} 
& \multicolumn{2}{c}{\textbf{Random Retrieval}} \\
{\scriptsize Metric} 
& {\scriptsize MSE} & {\scriptsize MAE}
& {\scriptsize MSE} & {\scriptsize MAE}
& {\scriptsize MSE} & {\scriptsize MAE}
& {\scriptsize MSE} & {\scriptsize MAE} \\
\midrule
ETTh1        & \textbf{0.415} & \textbf{0.435} & 0.421 & 0.435 & 1.328 & 0.740 & 0.421 & 0.437 \\
ETTm1        & \textbf{0.346} & \textbf{0.378} & 0.359 & 0.381 & 1.272 & 0.699 & 0.361 & 0.383 \\
Exchange  & 0.394 & 0.415  & 0.411 & 0.420 & \textbf{0.377} & \textbf{0.409}  & 0.402 & 0.416  \\
Electricity  & \textbf{0.152} & \textbf{0.251} & 0.163 & 0.257 & 1.613 & 0.959 & 0.164 & 0.257 \\
Traffic      & \textbf{0.395} & \textbf{0.279} & 0.414 & 0.287 & 2.768 & 1.090 & 0.414 & 0.287 \\
\bottomrule
\end{tabular}%
}
\end{table}

\begin{table}[]
\centering
\small
\caption{Ablation study of components in the retriever. We report the average MSE and MAE across four forecasting horizons. The best result in each row is highlighted in bold.}
\label{tab:saraf_ablation_components}
\resizebox{\columnwidth}{!}{%
\begin{tabular}{lcccccccccc}
\hline
\textbf{Models} & \multicolumn{2}{c}{\textbf{SARAF}} & \multicolumn{2}{c}{\textbf{w/o time}} & \multicolumn{2}{c}{\textbf{w/o div}} & \multicolumn{2}{c}{\textbf{w/o sta}} & \multicolumn{2}{c}{\textbf{w/o div+sta}} \\
\scriptsize Metric & \scriptsize MSE & \scriptsize MAE & \scriptsize MSE & \scriptsize MAE & \scriptsize MSE & \scriptsize MAE & \scriptsize MSE & \scriptsize MAE & \scriptsize MSE & \scriptsize MAE \\ \hline
ETTh1 & \textbf{0.415} & \textbf{0.435} & 0.420 & 0.437 & 0.415 & 0.435 & 0.416 & 0.435 & 0.416 & 0.435 \\
ETTm1 & \textbf{0.346} & 0.378 & 0.347 & \textbf{0.378} & 0.347 & 0.378 & 0.348 & 0.379 & 0.348 & 0.379\\
Exchange & \textbf{0.394} & \textbf{0.415} & 0.398 & 0.421 & 0.400 & 0.418 & 0.398 & 0.416 & 0.399 & 0.417 \\
Electricity & 0.152 & \textbf{0.251} & 0.156 & 0.254 & \textbf{0.151} & 0.251 & 0.152 & 0.252 & 0.151 & 0.251 \\
Traffic & \textbf{0.395} & \textbf{0.279} & 0.402 & 0.281 & 0.396 & 0.280 & 0.396 & 0.281 & 0.397 & 0.280 \\ \hline
\end{tabular}%
}
\end{table}

\section{Experiment}
\subsection{Experimental Settings}

\textbf{Datasets.}
We evaluate SARAF on eight widely used multivariate time-series benchmarks: ETTh1, ETTh2, ETTm1, ETTm2, Exchange, Solar, Electricity, and Traffic~\cite{wu2021autoformer}. These datasets cover diverse channel dimensions, sampling frequencies, and degrees of stationarity. As summarized in Table~\ref{tab:dataset_stats_simple}, we report the number of channels, input length, forecasting horizons, dataset size, split ratio, and sampling frequency for each dataset. To further characterize their temporal dynamics, we provide two stationarity-related statistics: the proposed stationarity score and the ADF stationary ratio. A higher stationarity score indicates more stationary temporal patterns, while a lower score suggests stronger non-stationarity. The ADF stationary ratio denotes the proportion of channels identified as stationary by the ADF test. A more detailed comparison is provided in Appendix~\ref{app:stationarity_score}, where the proposed stationarity score shows a broadly consistent trend with the ADF-based stationary ratio.

\begin{table*}[]
\centering
\caption{Multivariate time series forecasting results with a look-back window of 720 and horizons $H\in\{96,192,336,720\}$. Best metric is bold and second-best is underlined.}
\small
\setlength{\tabcolsep}{1mm}
\resizebox{0.9\textwidth}{!}{%
\begin{tabular}{cc|cc|cc|cc|cc|cc|cc|cc|cc|cc|cc}
\toprule
\multicolumn{2}{c|}{Models} & \multicolumn{2}{c|}{SARAF (ours)} & \multicolumn{2}{c|}{RAFT} & \multicolumn{2}{c|}{DUET} & \multicolumn{2}{c|}{TimeMixer} & \multicolumn{2}{c|}{CycleNet} & \multicolumn{2}{c|}{PatchTST} & \multicolumn{2}{c|}{DLinear} & \multicolumn{2}{c|}{TimesNet} & \multicolumn{2}{c|}{Stationary} & \multicolumn{2}{c}{Autoformer} \\ \midrule
\multicolumn{2}{c|}{Metric} & MSE & MAE & MSE & MAE & MSE & \multicolumn{1}{c|}{MAE} & MSE & MAE & MSE & \multicolumn{1}{c|}{MAE} & MSE & MAE & MSE & \multicolumn{1}{c|}{MAE} & MSE & MAE & MSE & \multicolumn{1}{c|}{MAE} & MSE & MAE \\ \midrule
\multicolumn{1}{c|}{\multirow{5}{*}{\rotatebox{90}{ETTh1}}} & 96 & \underline{0.371} & \underline{0.399} & \textbf{0.366} & \textbf{0.396} & 0.382 & 0.407 & 0.408 & 0.429 & 0.396 & 0.418 & 0.539 & 0.513 & 0.418 & 0.444 & 0.858 & 0.667 & 0.553 & 0.536 & 0.533 & 0.522 \\
\multicolumn{1}{c|}{} & 192 & \underline{0.404} & \underline{0.420} & \textbf{0.403} & \textbf{0.419} & 0.421 & 0.439 & 0.435 & 0.444 & 0.435 & 0.444 & 0.611 & 0.551 & 0.461 & 0.471 & 0.915 & 0.683 & 0.509 & 0.507 & 0.580 & 0.545 \\
\multicolumn{1}{c|}{} & 336 & \textbf{0.435} & \underline{0.447} & \underline{0.437} & \textbf{0.443} & 0.440 & 0.453 & 0.464 & 0.462 & 0.468 & 0.465 & 0.675 & 0.582 & 0.527 & 0.506 & 0.892 & 0.672 & 0.564 & 0.553 & 0.627 & 0.575 \\
\multicolumn{1}{c|}{} & 720 & \textbf{0.451} & \textbf{0.473} & 0.467 & \underline{0.477} & 0.479 & 0.490 & \underline{0.463} & 0.477 & 0.473 & 0.487 & 0.716 & 0.616 & 0.680 & 0.612 & 0.971 & 0.693 & 0.741 & 0.605 & 0.660 & 0.582 \\ \cline{2-22} 
\multicolumn{1}{c|}{} & Avg & \textbf{0.415} & \underline{0.435} & \underline{0.418} & \textbf{0.434} & 0.430 & 0.447 & 0.442 & 0.453 & 0.443 & 0.454 & 0.635 & 0.565 & 0.521 & 0.508 & 0.909 & 0.679 & 0.592 & 0.550 & 0.600 & 0.556 \\ \hline
\multicolumn{1}{c|}{\multirow{5}{*}{\rotatebox{90}{ETTh2}}} & 96 & \textbf{0.271} & \textbf{0.335} & \underline{0.278} & 0.345 & 0.307 & 0.369 & 0.310 & 0.374 & 0.283 & \underline{0.344} & 0.385 & 0.428 & 0.391 & 0.436 & 0.438 & 0.463 & 0.399 & 0.444 & 0.328 & 0.387 \\
\multicolumn{1}{c|}{} & 192 & \textbf{0.331} & \textbf{0.378} & \underline{0.347} & 0.393 & 0.352 & 0.397 & 0.353 & 0.398 & 0.350 & \underline{0.391} & 0.408 & 0.443 & 0.447 & 0.469 & 0.477 & 0.486 & 0.416 & 0.455 & 0.378 & 0.419 \\
\multicolumn{1}{c|}{} & 336 & \textbf{0.359} & \textbf{0.408} & \underline{0.374} & 0.423 & 0.375 & 0.416 & 0.381 & 0.418 & 0.377 & \underline{0.415} & 0.422 & 0.451 & 0.422 & 0.454 & 0.507 & 0.503 & 0.439 & 0.473 & 0.391 & 0.432 \\
\multicolumn{1}{c|}{} & 720 & \textbf{0.399} & 0.447 & 0.432 & 0.472 & \underline{0.400} & \textbf{0.438} & 0.433 & 0.456 & 0.401 & \underline{0.440} & 0.434 & 0.465 & 0.519 & 0.527 & 0.521 & 0.514 & 0.537 & 0.540 & 0.457 & 0.476 \\ \cline{2-22} 
\multicolumn{1}{c|}{} & Avg & \textbf{0.340} & \textbf{0.392} & 0.358 & 0.408 & 0.358 & 0.405 & 0.369 & 0.412 & \underline{0.353} & \underline{0.398} & 0.412 & 0.446 & 0.445 & 0.472 & 0.486 & 0.491 & 0.448 & 0.478 & 0.388 & 0.428 \\ \hline
\multicolumn{1}{c|}{\multirow{5}{*}{\rotatebox{90}{ETTm1}}} & 96 & \textbf{0.297} & \textbf{0.348} & \underline{0.303} & \underline{0.350} & 0.297 & 0.351 & 0.326 & 0.369 & 0.323 & 0.363 & 0.523 & 0.487 & 0.352 & 0.391 & 0.857 & 0.635 & 0.517 & 0.489 & 0.412 & 0.423 \\
\multicolumn{1}{c|}{} & 192 & 0.331 & \underline{0.370} & \textbf{0.327} & \textbf{0.365} & \underline{0.330} & 0.371 & 0.358 & 0.389 & 0.349 & 0.378 & 0.479 & 0.453 & 0.375 & 0.406 & 0.925 & 0.664 & 0.533 & 0.497 & 0.476 & 0.455 \\
\multicolumn{1}{c|}{} & 336 & \underline{0.357} & \underline{0.386} & \textbf{0.353} & \textbf{0.381} & 0.358 & 0.388 & 0.383 & 0.406 & 0.378 & 0.395 & 0.568 & 0.494 & 0.408 & 0.428 & 0.903 & 0.636 & 0.557 & 0.509 & 0.556 & 0.523 \\
\multicolumn{1}{c|}{} & 720 & \textbf{0.401} & \textbf{0.410} & \underline{0.406} & \underline{0.412} & 0.411 & 0.419 & 0.474 & 0.453 & 0.426 & 0.421 & 0.624 & 0.533 & 0.464 & 0.463 & 1.020 & 0.674 & 0.496 & 0.485 & 0.593 & 0.517 \\ \cline{2-22} 
\multicolumn{1}{c|}{} & Avg & \textbf{0.346} & \underline{0.378} & \underline{0.347} & \textbf{0.377} & 0.349 & 0.382 & 0.385 & 0.404 & 0.369 & 0.390 & 0.548 & 0.492 & 0.400 & 0.422 & 0.926 & 0.653 & 0.526 & 0.495 & 0.509 & 0.480 \\ \hline
\multicolumn{1}{c|}{\multirow{5}{*}{\rotatebox{90}{ETTm2}}} & 96 & \textbf{0.160} & \textbf{0.252} & \underline{0.164} & 0.256 & 0.165 & \underline{0.254} & 0.169 & 0.261 & 0.173 & 0.266 & 0.272 & 0.335 & 0.210 & 0.296 & 0.351 & 0.388 & 0.328 & 0.387 & 0.235 & 0.329 \\
\multicolumn{1}{c|}{} & 192 & \textbf{0.215} & \textbf{0.292} & \underline{0.220} & 0.296 & 0.223 & \underline{0.294} & 0.231 & 0.304 & 0.226 & 0.302 & 0.308 & 0.353 & 0.259 & 0.326 & 0.397 & 0.412 & 0.330 & 0.382 & 0.324 & 0.370 \\
\multicolumn{1}{c|}{} & 336 & \textbf{0.268} & \textbf{0.329} & \underline{0.275} & 0.335 & 0.279 & \underline{0.331} & 0.283 & 0.338 & 0.276 & 0.335 & 0.347 & 0.383 & 0.307 & 0.358 & 0.432 & 0.433 & 0.369 & 0.410 & 0.394 & 0.419 \\
\multicolumn{1}{c|}{} & 720 & \textbf{0.349} & \underline{0.383} & 0.369 & 0.398 & \underline{0.351} & \textbf{0.378} & 0.358 & 0.386 & 0.362 & 0.390 & 0.465 & 0.452 & 0.385 & 0.405 & 0.491 & 0.472 & 0.439 & 0.458 & 0.488 & 0.466 \\ \cline{2-22} 
\multicolumn{1}{c|}{} & Avg & \textbf{0.248} & \textbf{0.314} & 0.257 & \underline{0.321} & \underline{0.254} & 0.314 & 0.260 & 0.323 & 0.259 & 0.323 & 0.348 & 0.381 & 0.290 & 0.346 & 0.418 & 0.426 & 0.367 & 0.409 & 0.360 & 0.396 \\ \hline
\multicolumn{1}{c|}{\multirow{5}{*}{\rotatebox{90}{Exchange}}} & 96 & \textbf{0.085} & \textbf{0.202} & 0.090 & \underline{0.207} & 0.098 & 0.224 & 0.092 & 0.219 & 0.091 & 0.212 & 0.137 & 0.271 & \underline{0.088} & 0.211 & 0.233 & 0.334 & 0.688 & 0.644 & 0.237 & 0.356 \\
\multicolumn{1}{c|}{} & 192 & 0.186 & \textbf{0.304} & 0.201 & 0.317 & 0.206 & 0.329 & 0.205 & 0.326 & \underline{0.184} & \underline{0.305} & 0.209 & 0.332 & \textbf{0.177} & 0.307 & 0.410 & 0.434 & 0.855 & 0.724 & 0.393 & 0.470 \\
\multicolumn{1}{c|}{} & 336 & \textbf{0.349} & \textbf{0.426} & 0.377 & 0.445 & 0.371 & 0.446 & 0.370 & 0.442 & 0.361 & \underline{0.438} & \underline{0.351} & 0.440 & 0.369 & 0.456 & 0.612 & 0.543 & 1.000 & 0.785 & 0.618 & 0.596 \\
\multicolumn{1}{c|}{} & 720 & \underline{0.955} & \underline{0.728} & 1.129 & 0.805 & 1.202 & 0.783 & 0.996 & 0.746 & 1.043 & 0.756 & 1.046 & 0.778 & \textbf{0.911} & \textbf{0.713} & 1.469 & 0.825 & 1.807 & 1.048 & 1.910 & 1.006 \\ \cline{2-22} 
\multicolumn{1}{c|}{} & Avg & \underline{0.394} & \textbf{0.415} & 0.449 & 0.444 & 0.469 & 0.446 & 0.416 & 0.433 & 0.420 & 0.428 & 0.436 & 0.455 & \textbf{0.386} & \underline{0.422} & 0.681 & 0.534 & 1.087 & 0.800 & 0.789 & 0.607 \\ \hline
\multicolumn{1}{c|}{\multirow{5}{*}{\rotatebox{90}{Solar}}} & 96 & 0.182 & 0.244 & 0.188 & 0.239 & 0.183 & \underline{0.237} & 0.193 & 0.257 & 0.203 & 0.271 & \textbf{0.167} & \textbf{0.229} & 0.191 & 0.258 & \underline{0.174} & 0.260 & 0.623 & 0.594 & 0.178 & 0.229 \\
\multicolumn{1}{c|}{} & 192 & 0.204 & 0.254 & 0.206 & 0.252 & \underline{0.191} & \textbf{0.237} & 0.223 & 0.265 & 0.220 & 0.283 & \textbf{0.183} & \underline{0.244} & 0.211 & 0.273 & 0.194 & 0.273 & 0.805 & 0.629 & 0.208 & 0.279 \\
\multicolumn{1}{c|}{} & 336 & 0.216 & 0.263 & 0.220 & 0.261 & \underline{0.206} & \underline{0.255} & 0.234 & 0.275 & 0.232 & 0.294 & \textbf{0.197} & \textbf{0.253} & 0.228 & 0.288 & 0.222 & 0.289 & 0.701 & 0.627 & 0.210 & 0.259 \\
\multicolumn{1}{c|}{} & 720 & 0.224 & 0.270 & 0.228 & 0.270 & 0.213 & \textbf{0.253} & 0.239 & 0.292 & 0.241 & 0.287 & \textbf{0.208} & \underline{0.260} & 0.236 & 0.295 & 0.216 & 0.267 & 0.691 & 0.650 & \underline{0.211} & 0.276 \\ \cline{2-22} 
\multicolumn{1}{c|}{} & Avg & 0.206 & 0.258 & 0.211 & 0.255 & \underline{0.198} & \textbf{0.245} & 0.222 & 0.272 & 0.224 & 0.284 & \textbf{0.189} & \underline{0.246} & 0.216 & 0.278 & 0.201 & 0.272 & 0.705 & 0.625 & 0.202 & 0.261 \\ \hline
\multicolumn{1}{c|}{\multirow{5}{*}{\rotatebox{90}{Electricity}}} & 96 & 0.130 & 0.228 & 0.131 & 0.229 & 0.132 & \textbf{0.224} & 0.135 & 0.237 & \underline{0.130} & 0.226 & 0.131 & \underline{0.225} & \textbf{0.128} & 0.227 & 0.181 & 0.288 & 0.247 & 0.352 & 0.183 & 0.287 \\
\multicolumn{1}{c|}{} & 192 & \underline{0.143} & \underline{0.241} & 0.147 & 0.243 & 0.159 & 0.248 & 0.156 & 0.256 & 0.145 & \textbf{0.240} & 0.147 & 0.242 & \textbf{0.141} & 0.241 & 0.193 & 0.296 & 0.234 & 0.342 & 0.192 & 0.296 \\
\multicolumn{1}{c|}{} & 336 & \underline{0.160} & 0.260 & 0.160 & 0.257 & 0.168 & 0.260 & 0.170 & 0.271 & 0.161 & \textbf{0.257} & 0.163 & \underline{0.258} & \textbf{0.158} & 0.259 & 0.205 & 0.305 & 0.229 & 0.337 & 0.197 & 0.299 \\
\multicolumn{1}{c|}{} & 720 & \textbf{0.173} & \textbf{0.274} & \underline{0.187} & \underline{0.282} & 0.192 & 0.283 & 0.215 & 0.311 & 0.200 & 0.291 & 0.208 & 0.299 & 0.194 & 0.293 & 0.232 & 0.326 & 0.255 & 0.359 & 0.217 & 0.319 \\ \cline{2-22} 
\multicolumn{1}{c|}{} & Avg & \textbf{0.152} & \textbf{0.251} & 0.156 & \underline{0.253} & 0.163 & 0.254 & 0.169 & 0.269 & 0.159 & 0.254 & 0.162 & 0.256 & \underline{0.155} & 0.255 & 0.203 & 0.304 & 0.241 & 0.348 & 0.197 & 0.300 \\ \hline
\multicolumn{1}{c|}{\multirow{5}{*}{\rotatebox{90}{Traffic}}} & 96 & 0.371 & 0.267 & 0.376 & 0.269 & \textbf{0.346} & \textbf{0.232} & 0.462 & 0.285 & 0.386 & 0.273 & \underline{0.367} & \underline{0.251} & 0.379 & 0.266 & 0.601 & 0.322 & 0.654 & 0.402 & 0.626 & 0.345 \\
\multicolumn{1}{c|}{} & 192 & \underline{0.385} & 0.274 & 0.391 & 0.277 & \textbf{0.365} & \textbf{0.241} & 0.473 & 0.296 & 0.398 & 0.276 & 0.387 & 0.278 & 0.389 & \underline{0.271} & 0.604 & 0.324 & 0.665 & 0.410 & 0.637 & 0.353 \\
\multicolumn{1}{c|}{} & 336 & \underline{0.398} & 0.282 & 0.402 & 0.281 & \textbf{0.380} & \textbf{0.248} & 0.498 & 0.296 & 0.412 & 0.284 & 0.398 & \underline{0.271} & 0.403 & 0.279 & 0.626 & 0.337 & 0.707 & 0.445 & 0.641 & 0.356 \\
\multicolumn{1}{c|}{} & 720 & \textbf{0.425} & 0.295 & 0.435 & 0.298 & \underline{0.428} & \textbf{0.273} & 0.506 & 0.313 & 0.450 & 0.305 & 0.435 & \underline{0.291} & 0.450 & 0.305 & 0.666 & 0.351 & 0.689 & 0.422 & 0.659 & 0.360 \\ \cline{2-22} 
\multicolumn{1}{c|}{} & Avg & \underline{0.395} & 0.280 & 0.401 & 0.281 & \textbf{0.380} & \textbf{0.249} & 0.485 & 0.298 & 0.412 & 0.284 & 0.397 & \underline{0.273} & 0.405 & 0.280 & 0.624 & 0.334 & 0.679 & 0.420 & 0.641 & 0.353 \\ \hline
\end{tabular}%
}\label{main_result}
\end{table*}

\noindent\textbf{Baselines.} We compare SARAF against nine state-of-the-art forecasters. Autoformer~\cite{wu2021autoformer}, Non-stationary Transformer~\cite{liu2022non}, and PatchTST~\cite{nietime} are Transformer-based models. DLinear~\cite{zeng2023transformers} is a strong linear baseline with decomposition. RAFT~\cite{hanretrieval} augments forecasting with similarity-based multi-period retrieval. CycleNet~\cite{lin2024cyclenet} models periodicity on a linear backbone. TimesNet~\cite{wutimesnet} captures dominant periods via Fourier analysis. TimeMixer~\cite{wang2024timemixer} is a forecaster with multi-scale mixing. DUET~\cite{qiu2025duet} improves robustness via dual clustering over temporal patterns and cross-variable relations.

\noindent\textbf{Implementation details.}
For SARAF, we train the model for 10 epochs with a batch size of 32. The look-back window length is set to 720, and we evaluate four forecasting horizons, i.e., $H\in\{96,192,336,720\}$. Detailed hyperparameter settings for SARAF are provided in Appendix~\ref{hyperpara}. The retrieval database is constructed exclusively from the training split, ensuring that no information from the validation or test sets is used during retrieval. For a fair comparison, all baselines are evaluated under a unified protocol with the same data splits, look-back length, forecasting horizons, and evaluation metrics. Due to GPU memory constraints, we adjust the batch size for several baselines when required, following their original implementations, while keeping other experimental settings consistent whenever possible. Following prior work~\cite{hanretrieval}, we report Mean Squared Error (MSE) and Mean Absolute Error (MAE) as the evaluation metrics. All experiments are conducted on a single NVIDIA GPU, and the results reported in the tables are averaged over three independent runs.

\subsection{Main Results}
Table~\ref{main_result} summarizes the forecasting performance across all horizons and datasets.
Overall, SARAF achieves the best average MSE and MAE on 5 out of 8 datasets, showing competitive overall performance among the evaluated methods. When averaged across all datasets, SARAF reduces the MSE by 3.85\% and the MAE by 1.87\% compared with the retrieval-based forecaster RAFT. Compared with the strong forecasting baseline DUET, SARAF further achieves an average reduction of 4.05\% in MSE and 0.75\% in MAE. These results suggest that SARAF can provide useful gains over both retrieval-based and non-retrieval forecasting baselines, although the improvements vary across datasets and metrics. In particular, SARAF performs favorably on several datasets with different stationarity characteristics, including non-stationary series such as Exchange and more stationary benchmarks such as Electricity.

\subsection{Ablation Study}
\subsubsection{Ablation Study of the Retrieval Mechanism}
To evaluate the retrieval mechanism, we compare SARAF with three variants: (i) \textit{w/o Retriever}, which removes the retriever and relies solely on the forecaster; (ii) \textit{w/o Forecaster}, which removes the forecaster and predicts only by aggregating retrieved future segments; and (iii) \textit{Random Retrieval}, which keeps the fusion pipeline unchanged but replaces structured retrieval with randomly selected candidates. As shown in Table~\ref{tab:saraf_retrieval_ablation}, SARAF consistently outperforms \textit{w/o Retriever} in MSE and improves or matches MAE, confirming that retrieval provides useful complementary information beyond the forecaster. Meanwhile, \textit{Random Retrieval} is consistently worse than SARAF, showing that the gain comes from effective retrieval rather than retrieval alone. The \textit{w/o Forecaster} variant performs much worse on most datasets, indicating that retrieved futures generally require the forecaster to calibrate noise and mismatches. Notably, on Exchange, although removing the retriever still degrades performance and random retrieval remains inferior, \textit{w/o Forecaster} achieves the best results. This suggests that for highly non-stationary series, the retrieval branch itself can be especially informative, supporting our claim that retrieval becomes more important under stronger non-stationarity.

\subsubsection{Ablation Study of Components in Retriever}

We further conduct an ablation study on the key components of the retriever to identify which design choices contribute to SARAF's performance. In addition to the full model, we evaluate four variants: (i) \textit{w/o time}, which removes the time-aligned enhancement and treats all timestamps equally during retrieval; (ii) \textit{w/o div}, which disables diversity-aware retrieval and degenerates to a purely similarity-based Top-$K$ scheme; (iii) \textit{w/o sta}, which removes dataset-level stationarity estimation and uses fixed hyperparameters $\sigma=0.1$ and $\lambda=0.5$; and (iv) \textit{w/o div+sta}, which removes both stationarity estimation and diversity-aware selection, and uses a fixed $\sigma=0.1$.

As shown in Table~\ref{tab:saraf_ablation_components}, SARAF achieves the best or tied-best results on most metrics across the five datasets, confirming the overall effectiveness of the proposed retriever design. Among all components, the time-aligned enhancement shows the most consistent contribution: removing it increases MSE on all datasets and worsens or matches MAE, indicating that temporally aligned candidates are generally more informative for forecasting. The effects of diversity-aware retrieval and stationarity estimation are more dataset-dependent. On Exchange, where non-stationarity is stronger, removing diversity, stationarity awareness, or both leads to clear performance drops, suggesting that adaptive diversity control is important for covering heterogeneous temporal regimes and avoiding redundant neighbors. Similar but milder trends are observed on Traffic, while the differences on ETTm1 are relatively small. On relatively stable cases such as ETTh1 and Electricity, the differences are smaller, and Electricity even shows a slightly lower MSE after removing diversity-related components, while MAE remains tied. This is consistent with our design: when historical patterns are more stable, SARAF assigns less emphasis to diversification and relies more on similarity-based evidence. Overall, the ablation results indicate that time-aligned retrieval is broadly beneficial, while stationarity-conditioned diversity control mainly improves retrieval reliability when non-stationarity makes similarity-only neighbors more redundant or less dependable.

\begin{figure}[t]
    \centering
    \includegraphics[width=1\linewidth]{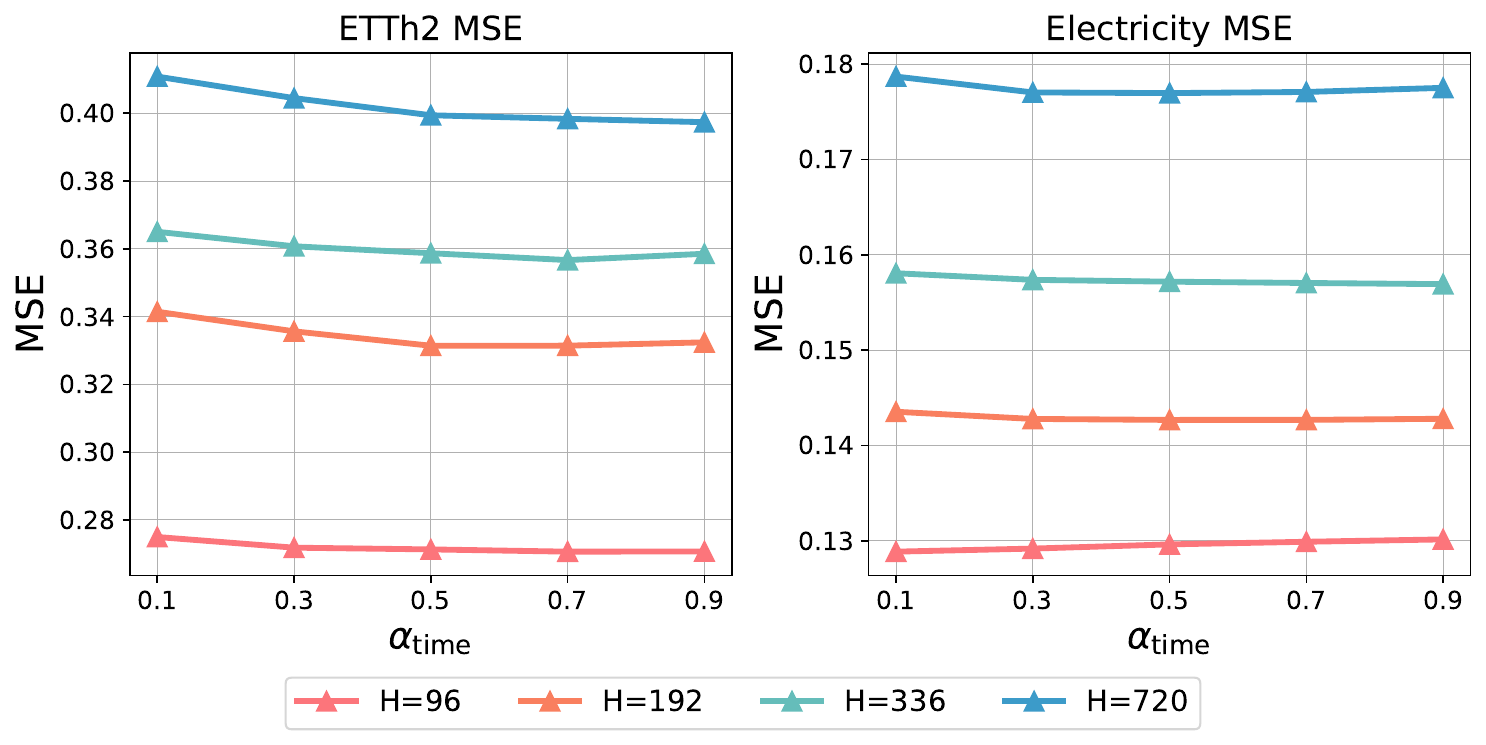}
    \caption{Hyperparameter sensitivity of $\alpha_{\text{time}}$ on ETTh2 and Electricity in terms of MSE across four horizons.}
    \label{fig:time} 
\end{figure}

\begin{figure}[t]
    \centering
    \includegraphics[width=1\linewidth]{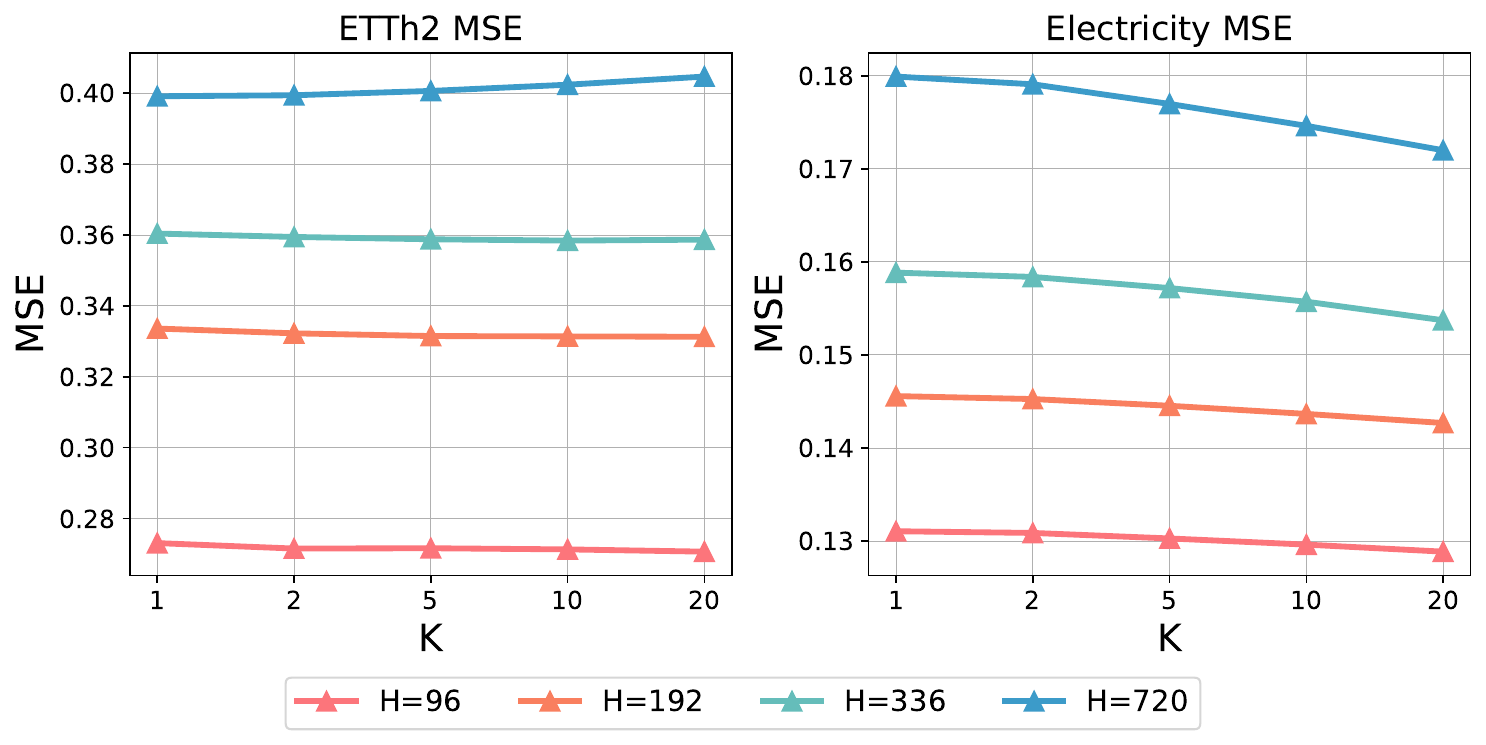}
    \caption{Hyperparameter sensitivity of $K$ on ETTh2 and Electricity in terms of MSE across four horizons.}
    \label{fig:topk} 
\end{figure}

\begin{figure}[t]
    \centering
    \includegraphics[width=1\linewidth]{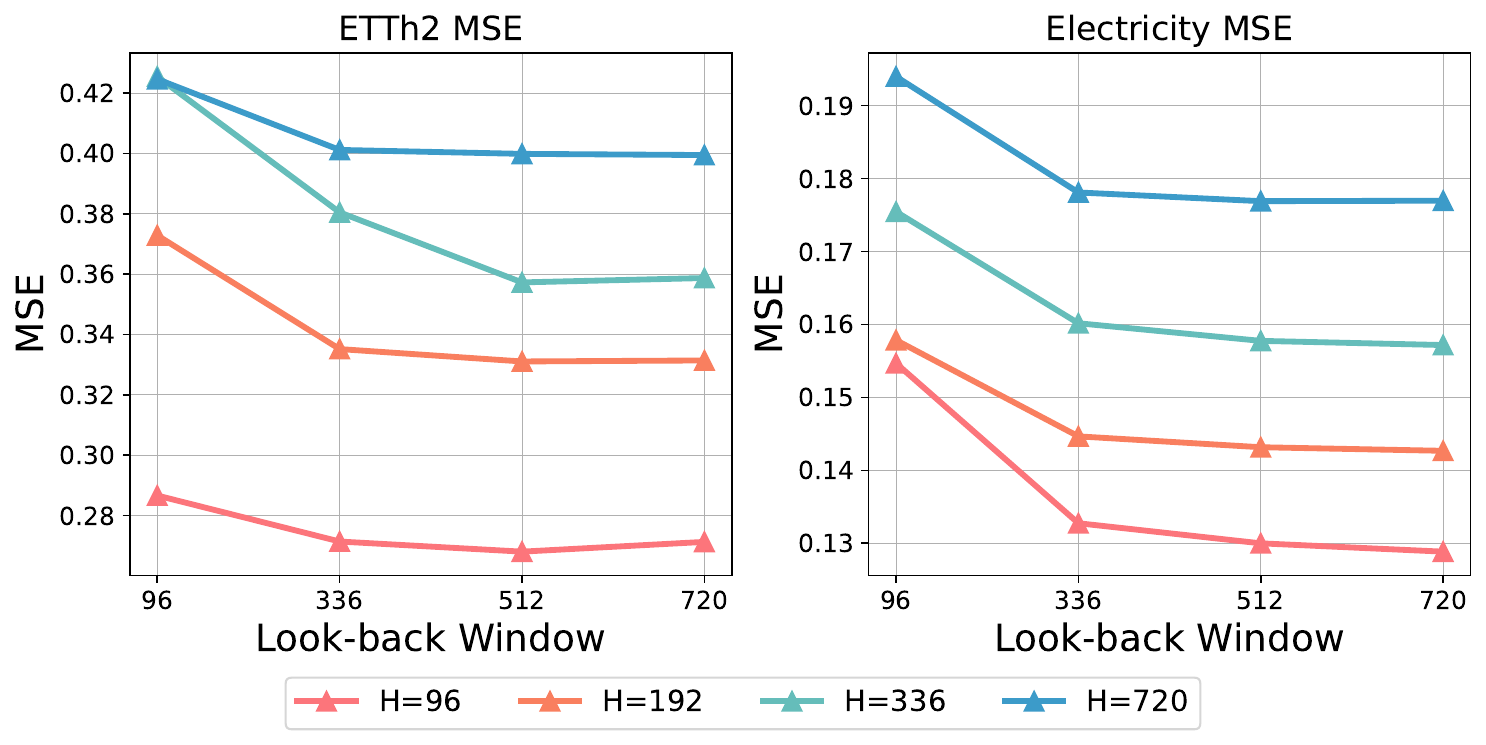}
    \caption{Hyperparameter sensitivity of look-back window on ETTh2 and Electricity in terms of MSE across four horizons.}
    \label{fig:window} 
\end{figure}

\subsection{Hyperparameter Sensitivity Analysis}
\subsubsection{Time-Aligned Enhancement Weight}
$\alpha_{\text{time}}$ controls the contribution of time-aligned enhancement. We analyze the varying $\alpha_{\text{time}}$ on the ETTh2 and Electricity datasets. Fig.~\ref{fig:time} suggests that the effect of the time-alignment weight $\alpha_{\text{time}}$ is dataset-dependent. On ETTh2, MSE changes more visibly as $\alpha_{\text{time}}$ varies, and the long-horizon case (H=720) shows the clearest drop when increasing $\alpha_{\text{time}}$ from low to moderate values, indicating that time alignment can matter more when forecasting further ahead. In contrast, Electricity is more stable across $\alpha_{\text{time}}$, with only small fluctuations across all horizons, consistent with a more stationary dataset where temporal alignment is less critical and performance is more robust to $\alpha_{\text{time}}$.

\subsubsection{Final Retrieval Set Size $K$} 
The sensitivity analysis of $K$ shows clear dataset-dependent behavior in Fig.~\ref{fig:topk}. Here, $K$ denotes the number of retrieved neighbors ultimately retained for fusion in the retrieval-augmented forecast. On Electricity, increasing $K$ consistently reduces MSE across all horizons, indicating that this relatively stationary dataset benefits from a broader retrieval context: aggregating more related segments improves the fused signal without introducing noticeable noise. In contrast, on ETTh2, the trend is much weaker and can even reverse at longer horizons. While small $K$ achieves comparable performance, overly large $K$ is more likely to admit less relevant candidates, which dilutes informative evidence and leads to higher error. Overall, these results suggest that the optimal retrieval budget is not universal and should be tuned to the dataset’s dynamics.

\subsubsection{Varying Look-back Window}
The look-back window length plays a critical role in time series forecasting, as it determines how much historical context is available for extrapolating future dynamics.
In Fig.~\ref{fig:window}, increasing window length on both ETTh2 and Electricity generally reduces MSE, and the largest gains arise when moving from short histories to moderately long windows.
The improvement is more pronounced at longer horizons, suggesting that additional context is particularly useful for capturing slow-varying trends and long-range temporal dependencies.
When the look-back window becomes very large, the gains tend to saturate and may slightly regress, reflecting diminishing returns and the increased chance of introducing less relevant history.

Beyond accuracy trends, a sufficiently long input is also important for retrieval-based forecasting.
Since retrieval relies on input similarity to identify historical segments with matching futures, the retrieval difficulty increases as the forecasting horizon grows.
If the look-back window is too short while the target horizon is long, input similarity becomes a weaker proxy for future similarity, making it substantially harder to retrieve truly informative analogues.
This motivates using a relatively long window in practice; in our setting, a look-back window of 720 steps remains a strong and stable choice, as it better matches long-horizon prediction and supports more reliable similarity-based retrieval.

\begin{figure}[t]
    \centering
    \includegraphics[width=1\linewidth]{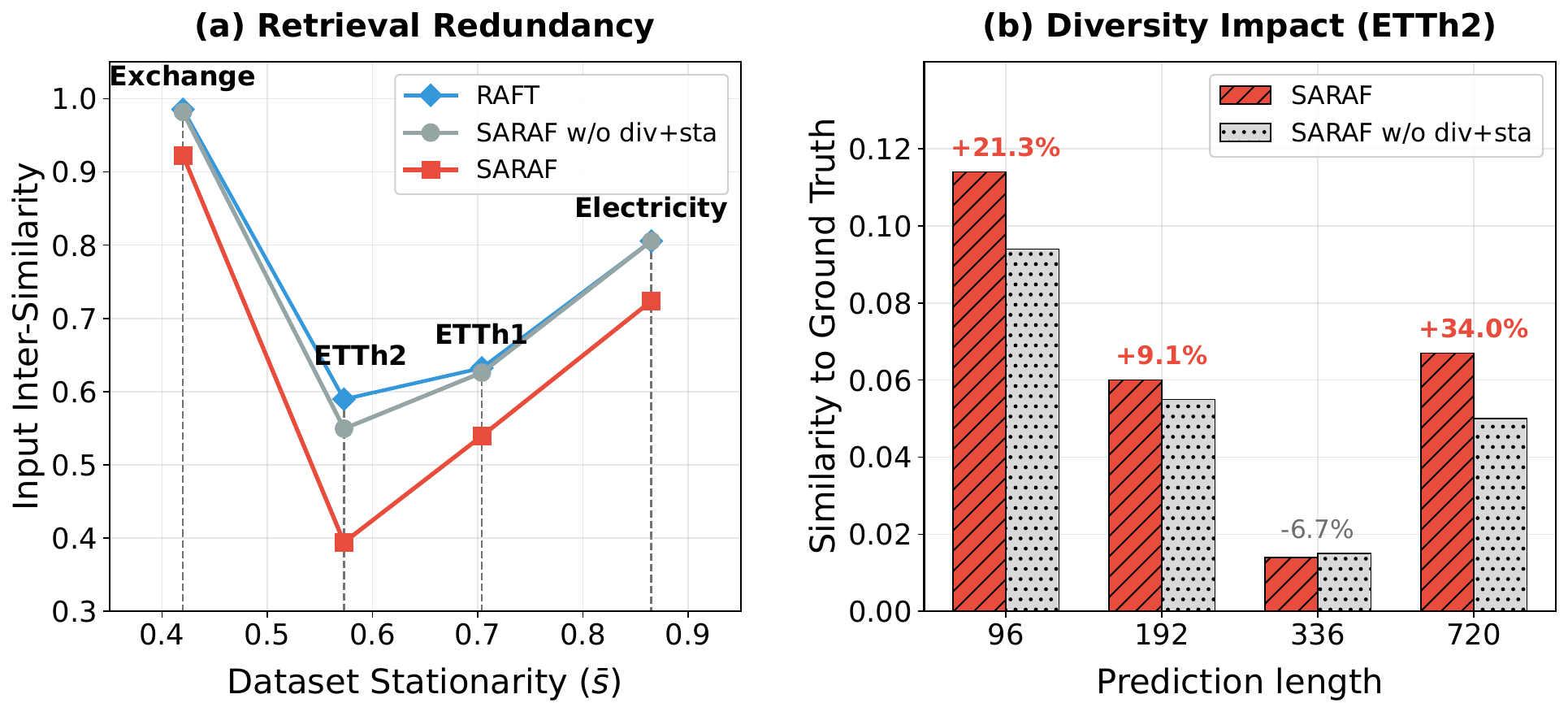}
    \caption{Diversity-Based Retrieval Analysis: (a) Retrieval redundancy across datasets; (b) Diversity-based retrieval impact on similarity between fused futures and ground truth.}
    \label{fig:diversity} 
\end{figure}

\subsection{Model Analysis}
\subsubsection{Diversity-Based Retrieval Analysis}
A key challenge in retrieval-augmented forecasting is that the retrieved evidence must be both \emph{relevant} and \emph{non-redundant}. To understand whether  diversity-aware design in SARAF truly improves evidence quality, we analyze the retrieved set from two perspectives: (i) redundancy within retrieved input windows, and (ii) the usefulness of the fused retrieved futures in matching the ground-truth future. Fig.~\ref{fig:diversity} (a) provides structural evidence of this issue across four datasets with different stationarity: RAFT and the ablated variant (w/o div+sta) exhibit consistently higher input inter-similarity among retrieved input windows, indicating that their Top-$K$ evidence tends to be near-duplicates. In contrast, SARAF consistently reduces intra-set similarity across datasets with different stationarity levels, largely mitigating the redundancy issue; on average, it lowers inter-similarity by 14.57\% compared to the ablated variant and by 16.09\% compared to RAFT. This reduction in redundancy translates into more useful evidence for forecasting. Fig.~\ref{fig:diversity} (b) further shows that, on the relatively non-stationary ETTh2 (stationarity score $\bar{s}=0.573$; ADF stationary ratio: 57.1\%), SARAF achieves higher similarity between the fused retrieved futures and the ground-truth future for most prediction lengths. In such datasets, SARAF intentionally increases the diversification strength, retrieving a broader set of candidate regimes; during fusion, it also spreads the aggregation weights beyond the single top match, assigning relatively more weight to the remaining retrieved samples so that multiple plausible futures can contribute. This motivates our stationarity-aware control to balance relevance and diversity adaptively across datasets.

\subsubsection{Retriever as an Add-On Module}
As the forecaster is a general backbone in the SARAF pipeline, as shown in Fig.~\ref{SARAF}, we replace the default linear layer with PatchTST, a Transformer-based model, and DLinear, a linear-based model, and integrate both with our retriever in a plug-and-play manner. Table~\ref{tab:backbone_avg_retrieval} reports the average results across four forecasting horizons on ETTh1 and ETTm1 under the same experimental setting as in the main results. The retrieval-augmented variants consistently improve their backbones in both MSE and MAE, demonstrating that our retriever generalizes well and effectively boosts diverse forecasters.

\begin{table}[t]
\centering
\small
\caption{Average performance across four forecasting horizons comparing backbones with and without the retriever. The better value within each backbone pair is in bold.}
\label{tab:backbone_avg_retrieval}
\begin{tabular}{l c c c c}
\toprule
\textbf{Dataset} & \multicolumn{2}{c}{\textbf{ETTh1}} & \multicolumn{2}{c}{\textbf{ETTm1}} \\
{\scriptsize Metric}  & {\scriptsize MSE} & {\scriptsize MAE} & {\scriptsize MSE} & {\scriptsize MAE} \\
\midrule
PatchTST & 0.635 & 0.565 & 0.548 & 0.492 \\
PatchTST + Retriever & \textbf{0.540} & \textbf{0.527} & \textbf{0.477} & \textbf{0.471} \\
\midrule
DLinear & 0.521 & 0.508 & 0.400 & 0.422 \\
DLinear + Retriever & \textbf{0.413} & \textbf{0.431} & \textbf{0.353} & \textbf{0.381} \\
\bottomrule
\end{tabular}
\end{table}

\subsubsection{Efficiency Analysis}
We compare the inference-time efficiency of four models under the same setting on ETTh1, using an input length of 720 and a forecasting horizon of 96, with batch size 32. All models are trained for 10 epochs under the same protocol, and we report parameter count, model memory, overall GPU memory usage during inference, and average inference speed per iteration. As reported in Table~\ref{tab:efficiency}, SARAF achieves the second fastest inference (0.334 ms/iter) while remaining compact in both parameter count (0.088 M) and model memory (0.335\,MiB), indicating a lightweight inference pipeline. In contrast, retrieval-based methods such as RAFT are slower and incurs higher memory usage, which is expected since it performs attention over retrieved candidates during inference. We further observe that SARAF has higher overall memory usage than pure forecasters such as DLinear and DUET, as it additionally maintains an external retrieval database and related runtime buffers. Nevertheless, its memory usage remains lower than that of RAFT, benefiting from its retrieval mechanism.

\begin{table}[t]
\centering
\small
\caption{Inference-time efficiency on ETTh1 with input length 720, forecasting horizon 96, and batch size 32. We report model parameters, model memory, overall memory, and inference speed per iteration; best results are in bold.}

\label{tab:efficiency}
\resizebox{\columnwidth}{!}{%
\begin{tabular}{l c c c c}
\toprule
\textbf{Model} & \textbf{Params (M)} & \textbf{Model Mem. (MiB)} & \textbf{All Mem. (MiB)} & \textbf{Speed (ms/iter)} \\
\midrule
SARAF   & \textbf{0.088} & \textbf{0.335} & 54.077 & 0.334 \\
DUET    & 6.660 & 25.408 & 36.030 & 7.202 \\
RAFT    & 0.104 & 0.397 & 122.996 & 0.590 \\
DLinear & 0.138 & 0.528 & \textbf{20.460} & \textbf{0.271} \\

\bottomrule
\end{tabular}%
}
\end{table}

\section{Conclusion}
In this paper, we studied a key limitation of retrieval-augmented time series forecasting: highly similar historical inputs do not always lead to similar future trajectories. Our analysis shows that the reliability of similarity-based retrieval is closely related to dataset stationarity; under non-stationarity, top-ranked neighbors may be redundant or associated with mismatched futures. Motivated by this observation, we proposed SARAF, a stationarity-aware retrieval-augmented forecasting framework that combines time-aligned similarity retrieval, stationarity-controlled diversity selection, and adaptive Gaussian aggregation. Experiments on eight multivariate forecasting datasets, together with comprehensive analyses, show that SARAF achieves competitive overall performance and provides useful average gains over strong retrieval-based and non-retrieval baselines, although the benefits vary across datasets and metrics. These results suggest that stationarity can serve as a practical signal for controlling how retrieved evidence should be selected and fused, offering a more reliable alternative to purely similarity-based retrieval under heterogeneous temporal regimes.

\bibliographystyle{ACM-Reference-Format}
\balance
\bibliography{KDD}

\newpage

\section*{Appendix}
\appendix

\section{Dataset Statistics and Stationarity Analysis}
\label{app:stationarity_score}

We further analyze the stationarity characteristics of the evaluated datasets. As shown in Table~\ref{tab:dataset_stats_simple}, Exchange is the least stationary dataset, with the lowest stationarity score and ADF stationary ratio. ETTh2, and to a lesser extent ETTm2, exhibit moderate non-stationarity, lying between Exchange and the more stable benchmarks. In contrast, Solar, Electricity, and Traffic are among the most stationary datasets, with consistently high stationarity scores and ADF stationary ratios, while ETTh1 and ETTm1 also appear relatively stationary overall.

We additionally report the ADF-based stationary ratio, obtained by applying the Augmented Dickey--Fuller (ADF) test to each channel and computing the fraction of channels classified as stationary. Overall, this ratio shows a positive association with our stationarity score: datasets with lower ADF stationary ratios tend to have smaller stationarity scores, whereas datasets with higher ADF stationary ratios generally exhibit larger scores. This consistency suggests that our stationarity score captures stationarity-related dataset characteristics that are broadly aligned with classical statistical testing, while remaining directly usable for adaptive retrieval control in SARAF.

\section{Motivation: When Similar Inputs Do Not Guarantee Similar Futures}
\label{Motivation_app}

Empirically, we observe that high similarity in historical inputs does not always imply high similarity in the corresponding futures, revealing a key limitation of similarity-only retrieval. For each dataset, we retrieve the Top-10 neighbors ranked by input similarity, compute their future similarities to the ground truth, and measure the agreement between the input-induced and future-induced rankings using Spearman's $\rho$. Pearson correlation is used as the similarity metric.

\begin{figure*}[t]
    \centering
    \includegraphics[width=1\linewidth]{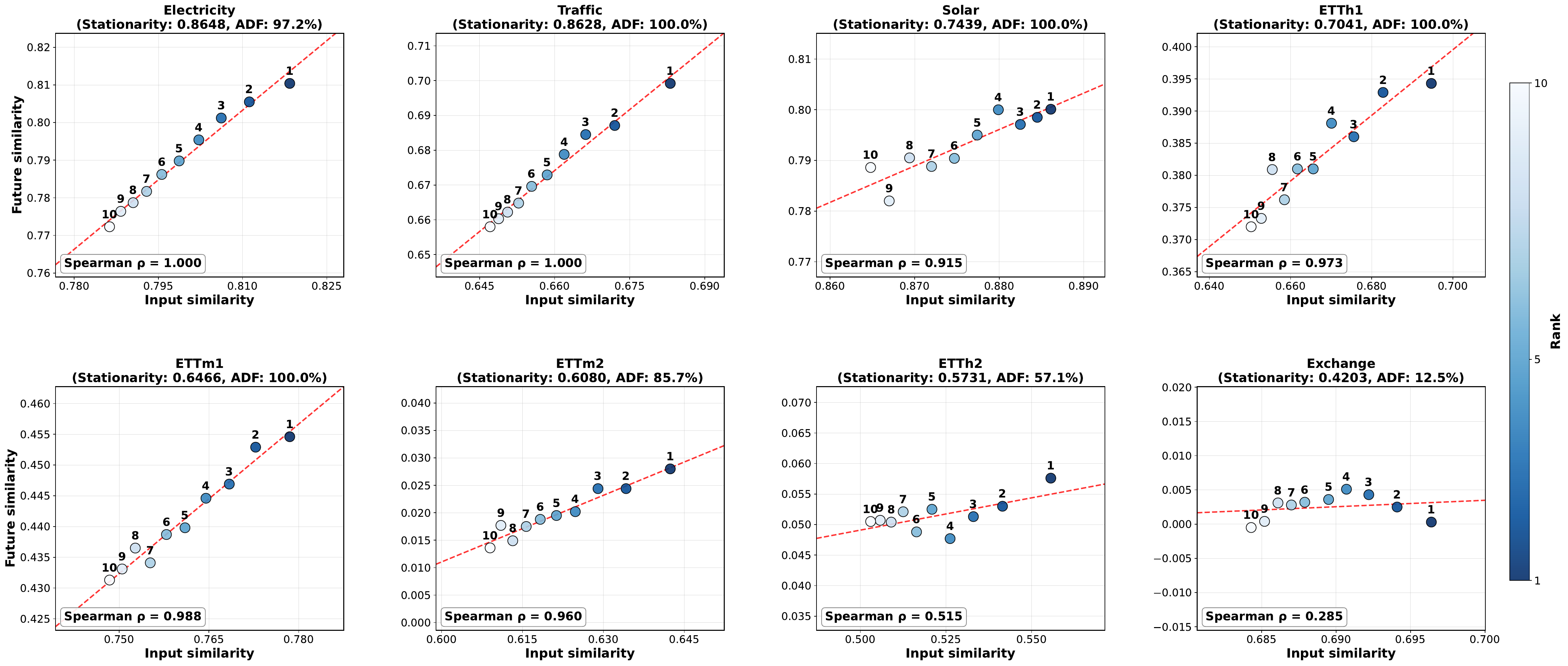}
    \caption{Motivation: Input similarity versus future similarity across eight datasets, ordered by stationarity (left-to-right/top-to-bottom from more to less stationary). Each point corresponds to the retrieved neighbor ranked by input similarity (rank labels and color). We report the Spearman rank correlation $\rho$ between the input-induced ranking and the future-similarity ranking in each subplot. More stationary datasets exhibit stronger rank agreement and higher future similarity, while non-stationary datasets show degraded alignment, highlighting reduced reliability of similarity-only retrieval.}
    \label{fig:motivation_scatter} 
\end{figure*}

As shown in Fig.~\ref{fig:motivation_scatter}, datasets with higher stationarity scores generally exhibit more reliable retrieval behavior. On highly stationary datasets such as Electricity and Traffic, the input-similarity ranking almost perfectly matches the future-similarity ranking, with Spearman $\rho=1.000$. Their retrieved futures also remain highly similar to the ground truth, indicating that historical similarity is a reliable proxy for future usefulness. In contrast, this alignment weakens substantially on less stationary datasets. For example, ETTh2 obtains a much lower rank correlation of $\rho=0.515$, while Exchange, the least stationary dataset, drops further to $\rho=0.285$ and exhibits future similarities concentrated near zero. These results show that similarity-only retrieval can become unreliable under strong non-stationarity or regime shifts.

\begin{figure}[t]
    \centering
    \includegraphics[width=1\linewidth]{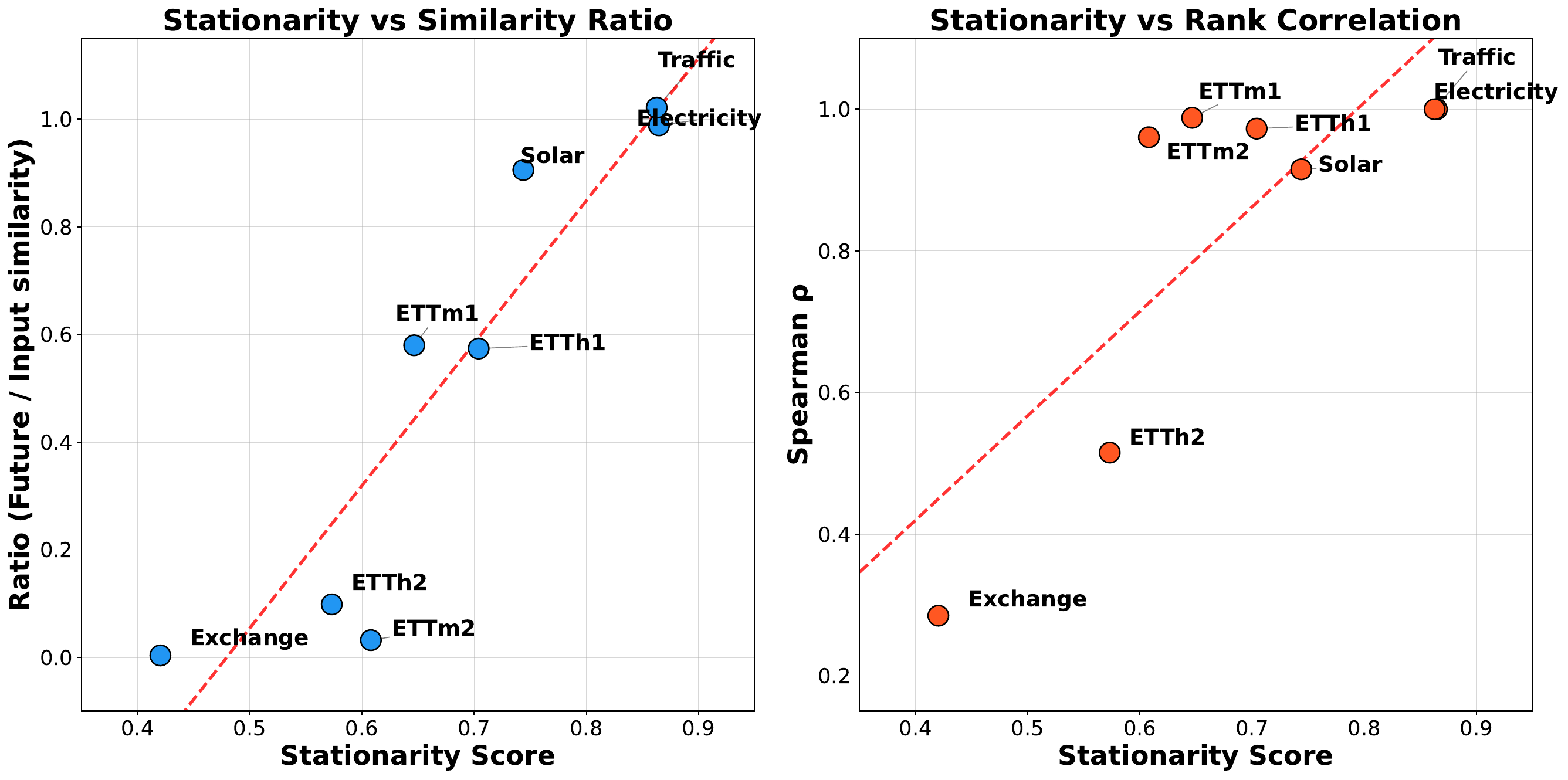}
    \caption{Retrieval reliability versus stationarity score. Left: future similarity divided by input similarity. Right: Spearman rank correlation between input similarity and future similarity rankings.}
    \label{fig:stationarity_vs_ratio_rho} 
\end{figure}

To further summarize this observation, Fig.~\ref{fig:stationarity_vs_ratio_rho} reports dataset-level retrieval reliability against the proposed stationarity score. We consider two indicators: the similarity retention ratio, defined as future similarity divided by input similarity, and the Spearman rank correlation between input- and future-induced rankings. Both indicators show a clear positive trend with stationarity: highly stationary datasets such as Electricity and Traffic achieve near-perfect rank agreement and high retention, whereas non-stationary datasets such as Exchange and ETTh2 show much lower reliability. This supports our motivation that stationarity is an important factor in determining when similarity-based retrieval is trustworthy, and motivates our stationarity-aware retrieval and fusion design.

\section{Complexity Analysis}
\label{sec:complexity}
\paragraph{Complexity of Stationarity Estimation.}
For each database window $x_i \in \mathbb{R}^{L\times C}$, we divide it into $W$ non-overlapping sub-windows and compute per-channel local means and standard deviations to obtain the instance-level stationarity score $\tilde{s}_i$ and the dataset-level score $\bar{s}$. Since these sub-windows jointly cover the original length-$L$ window, the local statistics require only a constant number of linear passes over $x_i$, costing $O(LC)$. Computing the variations $v_\mu$ and $v_\sigma$ over the $W$ local statistics further costs $O(WC)$. Thus, the per-window complexity is $O(LC+WC)$, and the total cost over $N$ windows is $O(NLC+NWC)$. As $W$ is fixed in practice and $d=LC$, this reduces to $O(Nd)$. The global scale term $\bar{\sigma}$ is also computed by a linear pass, so the overall complexity remains $O(Nd)$.

For comparison, applying the Augmented Dickey--Fuller (ADF) test~\cite{dickey1979distribution} independently to all $C$ channels requires running ADF on each full training sequence of length $T$. With automatic lag selection up to a maximum lag order $p_{\max}$, repeated lagged regression fitting gives an overall cost of $O(CTp_{\max}^{3})$. In contrast, our stationarity estimator operates directly on the sliding-window database $\mathbf{X}\in\mathbb{R}^{N\times L\times C}$ using moment-based reductions, yielding linear complexity $O(NLC)$. Therefore, it avoids repeated regression fitting and lag-order selection, and can be efficiently batched on GPUs.

\paragraph{Complexity of Time-aligned Similarity Retrieval.}
We compute conventional similarity scores $S_{\mathrm{sim}}(q,i)$ between the query and all $N$ database windows, which costs $O(Nd)$ time with $d=LC$.
The time-aligned bonus is computed from timestamps with constant-time operations per candidate, adding $O(N)$.
Selecting the Top-$M$ candidates from the fused scores costs $O(N\log M)$ via partial sorting, so the overall retrieval complexity is $O(Nd + N\log M)$.

\paragraph{Complexity of Stochastic MMR}
Our stochastic MMR selection over a Top-$M$ candidate pool runs in $O(MK)$ time to select Top-$K$ items (with incremental max updates). In contrast, the standard MMR formulation typically computes candidate-to-candidate redundancy via pairwise similarities, which costs $O(M^2 d)$ over the Top-$M$ pool, where $d=LC$ is the flattened window dimension.
By using a query-similarity--based redundancy proxy that reuses the computed $S_{\mathrm{sim}}(q,\cdot)$, we avoid quadratic pairwise similarity computation and achieve substantially lower complexity when $M\gg K$.

\begin{table*}[!t]
\centering
\small
\caption{Hyperparameter search ranges and fixed values for SARAF.}
\label{tab:saraf_hyperparams}
\begin{tabular}{l c c}
\toprule
\textbf{Hyperparameter} & \textbf{Description} & \textbf{Choices / Default} \\
\midrule
batch\_size      & Batch size for training                                   & 32 \\
seq\_len         & Look-back window length                                    & 720 \\
train\_epochs    & Training epochs                                             & 10 \\
learning\_rate   & Learning rate                                               & $\{0.01,\,0.001,\,0.0001\}$ \\
$W$              & Number of sub-windows in stationarity estimation            & 6 \\
$\sigma_{\min}$  & Minimum bandwidth of the Gaussian kernel                    & 0.05 \\
$\sigma_{\max}$  & Maximum bandwidth of the Gaussian kernel                    & 0.30 \\
$\lambda_{\min}$ & Minimum MMR balancing coefficient                           & 0.30 \\
$\lambda_{\max}$ & Maximum MMR balancing coefficient                           & 0.90 \\
$\alpha_{\text{time}}$ & Weight of the time-aligned enhancement                & $\{0.1,\,0.3,\,0.5,\,0.7,\,0.9\}$ \\
$M$              & Candidate pool size after similarity-based retrieval         & 100 \\
$K$              & Number of retrieved candidates after diversity-based retrieval & $\{1,\,2,\,3,\,5,\,10,\,20\}$ \\

\bottomrule
\end{tabular}
\end{table*}

\section{Hyperparameters}
\label{hyperpara}
This section summarizes the key hyperparameters used in SARAF over all datasets and the corresponding search ranges. Table~\ref{tab:saraf_hyperparams} reports the default settings and the candidate ranges adopted in our experiments. The exact tuned hyperparameter settings for each dataset are provided in our released code.

\section{Limitations}
\label{sec:limitation}
First, SARAF currently relies on a global similarity function over the multivariate window. A more fine-grained channel-wise or group-wise similarity could better capture heterogeneous dynamics across variables, but it would also increase retrieval and storage costs, making the accuracy--efficiency trade-off an important design choice.
Second, the retrieval database is built from dense sliding windows, which can be memory-intensive for long sequences and large datasets. Compressing the database through clustering, prototype selection, or learned indexing may reduce storage and latency while preserving most of the useful evidence.
Finally, our stationarity control is estimated at the dataset level, which may not fully reflect instance-level regime shifts. Extending the stationarity signal to be context-dependent is a promising direction for further improving adaptivity under non-stationarity.

\end{document}